\documentclass[runningheads]{llncs}

\usepackage{xcolor}
\definecolor{conceptred}{RGB}{220,50,47}
\definecolor{conceptblue}{RGB}{34,102,170}

\newcommand{\vistecicon}{%
\textsuperscript{\hspace{0.10em}\raisebox{-0.12ex}{\includegraphics[height=1.05ex]{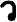}}}%
}

\newcommand{\chulaicon}{%
\textsuperscript{\hspace{0.10em}\raisebox{-0.12ex}{\includegraphics[height=1.05ex]{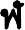}}}%
}

\newcommand{\vistecinsticon}{%
\textsuperscript{\hspace{0.10em}\raisebox{-0.15ex}{\includegraphics[height=1.24ex]{figures/icons/vistec2.pdf}}}%
}

\newcommand{\chulainsticon}{%
\textsuperscript{\hspace{0.10em}\raisebox{-0.15ex}{\includegraphics[height=1.30ex]{figures/icons/chula2.pdf}}}%
}
 
\usepackage{eccv}

\usepackage{eccvabbrv}

\usepackage{graphicx}
\usepackage{booktabs}
\usepackage{xcolor}
\usepackage[normalem]{ulem}
\usepackage{needspace}

\usepackage[accsupp]{axessibility}  

\usepackage[hidelinks]{hyperref}

\usepackage{orcidlink}
\DeclareMathOperator{\cossim}{sim}
\newcommand{\pushloss}{\ensuremath{\mathcal{L}_{\mathrm{push}}}}
\newcommand{\boldsubsection}[1]{\vspace{0.5em}\noindent\textbf{#1}}
\newcommand{\ourname}{Push–Pull Attentional Anchoring}
\newcommand{\ourshort}{PPAA}
\begin{document}

\title{\ourname{} for Diffusion Concept Erasure} 

\titlerunning{\ourname}

\author{
Nattanat Chatthee\vistecicon \and
Tagon Sompong\vistecicon \and
Ekapol Chuangsuwanich\chulaicon \and
Supasorn Suwajanakorn\vistecicon
}

\authorrunning{N.~Chatthee et al.}

\institute{
\vistecinsticon VISTEC, Thailand \quad \chulainsticon Chulalongkorn University, Thailand\\
\href{https://push-pull-concept-eraser.github.io}{\textcolor{blue!45!black}{\nolinkurl{push-pull-concept-eraser.github.io}}}
}







\maketitle

\vspace{-1.9em} 

\begin{figure}
    \centering
    \captionsetup{skip=5pt}
\includegraphics[width=1.00\linewidth]{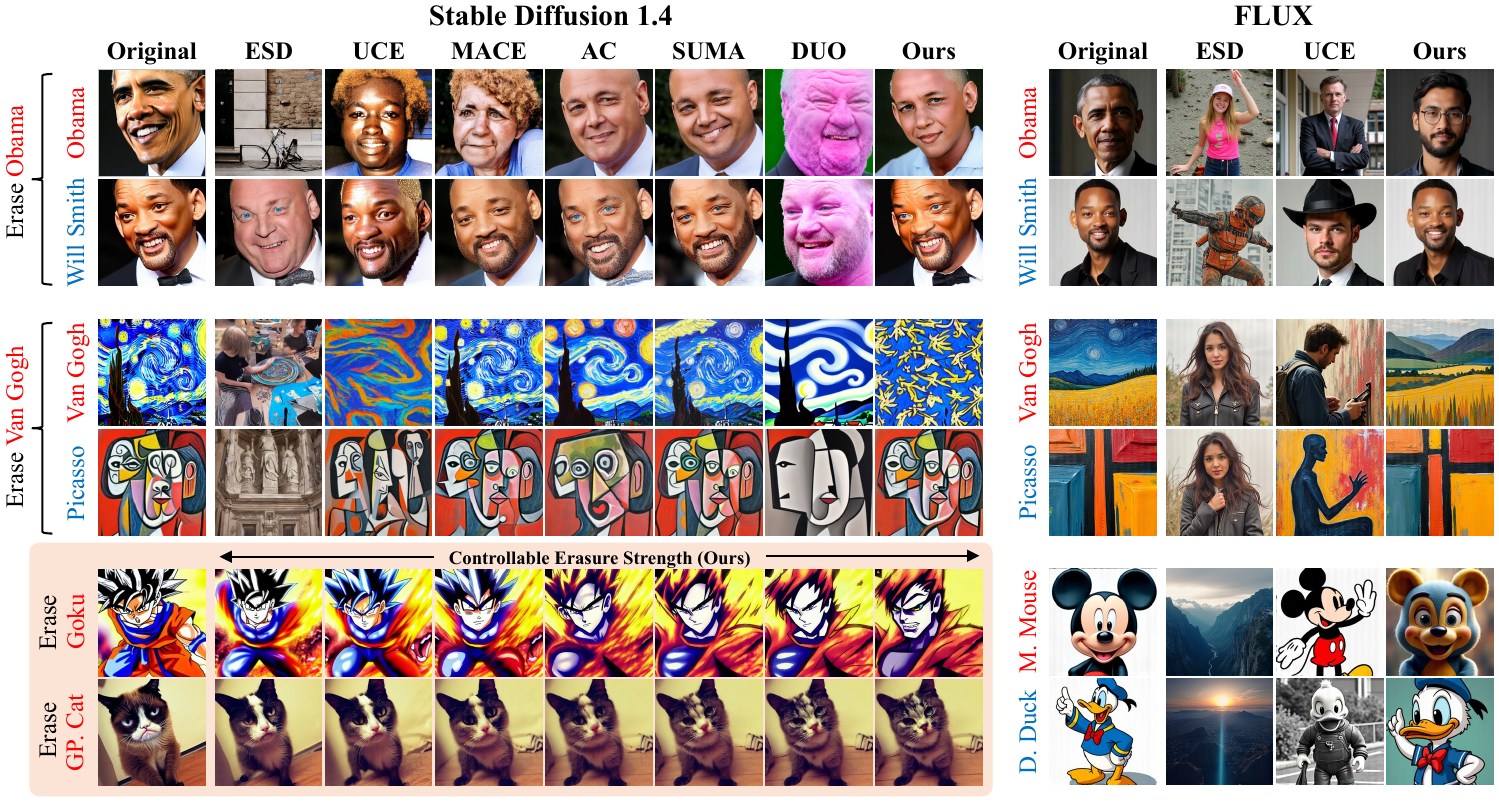}
\caption{Given a textual concept, our method erases the \textcolor{red}{target concept} while preserving \textcolor{conceptblue}{non-target concepts}, achieving a favorable erasure--preservation trade-off with controllable erasure strength and applicability to modern architectures such as FLUX.}
    \label{fig:teaser}
\end{figure}

\vspace{-2.50em}

\begin{abstract}

Rapid advances in diffusion models have raised concerns about privacy, harmful content, and copyright violations. Concept erasure methods aim to address these risks by removing undesirable concepts from pretrained models without full retraining.
However, existing approaches often degrade non-target generation quality through heuristic strength scaling (e.g., classifier-free guidance or negative prompt scaling), leading to uncontrolled and excessive semantic drift.
To address this, we propose \ourname{} (\ourshort), a mechanism in the cross-attention space that displaces target representations while anchoring the scale of erasure via bounded cosine-similarity constraints. By using a relative similarity ratio instead of an absolute difference, our adaptive threshold represents a meaningful percentage of semantic retention, facilitating a single value to be applied uniformly across layers.
We conduct extensive experiments across multiple concept categories, including celebrities, artistic styles, nudity, and intellectual property. Our method achieves favorable trade-offs against prior approaches on Stable Diffusion v1.4 and demonstrates its applicability to modern architectures such as FLUX.

\end{abstract}

\section{Introduction}
\label{sec:intro}
Rapid advances in text-to-image diffusion models have achieved impressive generative capabilities, but they have also raised critical concerns regarding privacy, harmful content, and copyright infringement. Concept erasure methods aim to mitigate these risks by selectively unlearning ``target’’ concepts from a pretrained diffusion model, avoiding the high cost of full model retraining. Driven by the discovery that early ``guidance-based’’ methods hide concepts rather than destroying them and leave models vulnerable to adversarial recovery~\cite{lu2025concepts}, recent state-of-the-art research has actively pursued ``destruction-based’’ removal to erase concepts as deeply and robustly as possible~\cite{nguyen2025suma, srivatsan2025stereo}. However, this growing focus on adversarial robustness often highlights a challenging trade-off with semantic preservation: the deeper the erasure, the higher the collateral damage to non-target generation quality.

To mitigate this trade-off, our study revisits the equally critical requirement of preserving non-target generation quality by examining the components of modern robust erasure methods. 
Recent state-of-the-art frameworks, such as STEREO~\cite{srivatsan2025stereo} and SUMA~\cite{nguyen2025suma}, employ an iterative two-stage design: (i) an adversarial discovery stage, which searches for prompts that can recover the erased concept, and (ii) an erasure stage, which updates the model weights to suppress the discovered targets. While Stage 1 effectively deters prompt-based attacks and improves robustness, existing Stage 2 erasure objectives tend to induce collateral damage. This largely stems from their reliance on uncontrolled parameter updates or rigid static mappings, directly inherited from earlier approaches~\cite{kumari2023ablating,gandikota2023erasing,lu2024mace,gandikota2024unified}, which excessively warp the local semantic neighborhood. Although existing methods can preserve broad, general-domain concepts (e.g., from MS-COCO~\cite{lin2014microsoft}), our study reveals that the more challenging preservation of closely related, same-domain peers, such as ``Will Smith'' when erasing ``Barack Obama,'' remains unsolved.

Since adversarial prompt discovery (Stage 1) functions as a distinct, orthogonal module, we focus on the erasure objective, e.g., those used in Stage 2~\cite{srivatsan2025stereo,nguyen2025suma} or as a standalone method~\cite{kumari2023ablating,gandikota2023erasing,lu2024mace,gandikota2024unified }, to isolate and resolve the preservation bottleneck. Following the standard setup~\cite{kumari2023ablating, gandikota2024unified,lu2025concepts}, which maps an erased concept to a user-provided generic ``anchor'' concept, we seek to finetune the model to suppress the conditional generation of the target concept. Our work prioritizes preserving closely related peer concepts, which are unavailable during optimization, with the goal of improving the overall erasure-preservation trade-off. 

To this end, we propose \ourname{} (\ourshort), a cross-attention mechanism that displaces target representations while explicitly anchoring the scale of erasure via bounded cosine-similarity constraints. Unlike objectives that rely on unbounded or rigid absolute-distance mappings~\cite{gandikota2024unified,lu2024mace,nguyen2025suma}, our mechanism introduces a margin that operates on the relative cosine ratio between the updated representation, its original embedding, and a generic reference. 
Because this relative formulation takes into account each layer's original target--anchor similarity, it yields an adaptive, concept-aware distance threshold that can be applied uniformly across layers. The resulting bounded push-pull objective enables sufficient target-concept displacement to evade target prompts while limiting unintended semantic modification. Furthermore, as our formulation can still serve as a Stage 2 erasure objective in the modern frameworks~\cite{srivatsan2025stereo,nguyen2025suma}, it can be combined with their adversarial discovery stage to further improve robustness.

We conduct extensive experiments across diverse concept categories, including celebrities, artistic styles, nudity, and intellectual property. Our evaluations explicitly measure both general-domain utility and strict same-domain preservation, demonstrating that \ourname{} achieves a favorable erasure-preservation trade-off compared to existing state-of-the-art methods.



\section{Related Work}
\subsection{Concept Erasure for Diffusion Models}

Early concept erasure methods~\cite{gandikota2023erasing, kumari2023ablating, lu2024mace, gandikota2024unified} focus on fine-tuning diffusion models to suppress the generation of specific target concepts by modifying the denoising process or its intermediate representations. ESD~\cite{gandikota2023erasing} explicitly steers the denoising trajectory of the target concept away from its original path by applying negative guidance~\cite{ho2022classifier, automatic1111_sd_webui, koulischer2024dynamic, chang2024contrastive }. In contrast, AC~\cite{kumari2023ablating} maps the denoising trajectory of the target concept to that of a more generic concept. MACE~\cite{lu2024mace} and UCE~\cite{gandikota2024unified} extend this idea by performing generic concept mapping in the key-value space of cross-attention layers rather than on the denoising trajectory. However, because the target concept and its mapped or negatively guided counterpart can be distant in representation space, as also observed in~\cite{nguyen2025suma}, these methods generally require substantial parameter updates to bridge this gap, which can significantly degrade the preservation of non-target concepts.

To address this limitation, SUMA~\cite{nguyen2025suma} leverages Textual Inversion (TI)~\cite{gal2022image} by using the final (late-step) TI embedding of the target concept and encouraging it to align with earlier-step TI embeddings obtained along the same TI optimization trajectory. These early-step embeddings are intended to capture more generic concept representations while remaining close to the target in embedding space. As a result, this alignment typically requires smaller weight modifications, leading to improved preservation of non-target concepts. However, using early-step TI embeddings as reference concepts may not be reliable in certain cases. In particular, these embeddings can rapidly converge toward the target concept itself, which limits erasure effectiveness~\cite{nguyen2025suma}.

Unlike approaches that rely on explicitly defined reference concepts for mapping, DUO~\cite{park2024direct} applies preference-based optimization~\cite{rafailov2023direct} within the diffusion denoising process. Because this optimization operates in the denoising space, which primarily captures low-level visual structure rather than high-level semantic representations~\cite{rombach2022high}, it encourages less accurate denoising for the target concept without explicitly enforcing semantic change. As a result, the optimization can produce outputs that are visually degraded yet semantically unchanged, leading to ineffective concept erasure.




\subsection{Adversarial Attacks for Concept Erasure}
Despite progress in concept erasure methods, recent adversarial attacks have successfully recovered supposedly erased concepts, highlighting the need for improved robustness. Recent studies~\cite{lu2025concepts, pham2023circumventing, zhang2024generate, rusanovsky2025memories, chin2023prompting4debugging, tsai2023ring} demonstrate that supposedly erased concepts can be recovered through adversarial attacks, exposing vulnerabilities in unlearned diffusion models. In particular,~\cite{lu2025concepts} provides a comprehensive evaluation of adversarial robustness using training-free probing techniques such as In-Painting Probing~\cite{abitbul2024much} and Partial Diffusion Completion~\cite{sharma2024unlearning}, which attempt to extract residual knowledge from erased models. These analyses reveal that many erasure approaches still retain traces of the removed concepts. Beyond probing-based attacks, several works~\cite{chin2023prompting4debugging, zhang2024generate, pham2023circumventing} introduce optimization-based attacks that search for adversarial prompts designed to recover the erased concept. These methods formulate prompt search as an optimization problem that minimizes the discrepancy between the denoising predictions of the erased model and those of the original model.
Although our method does not explicitly optimize for adversarial robustness, we evaluate our approach under existing adversarial attack protocols to assess its robustness.

\section{Proposed Method}
Our objective is to erase a target concept by directly updating model weights, rather than relying on easily bypassed guidance or auxiliary modules, while minimizing collateral damage to non-target concepts.
Determining the ideal replacement for an erasure target is inherently subjective; for example, erased ``Barack Obama'' could be mapped to ``Person,'' ``African American,'' or ``U.S. president,'' depending on the desired level of generality. Thus, following \cite{kumari2023ablating,gandikota2024unified,nguyen2025suma}, we assume that the generic replacement concept is provided by the user. However, non-target peer concepts (e.g., ``Will Smith'') that we intend to preserve remain unavailable during optimization and are reserved strictly for evaluation.

\subsection{Semantic Modification Constraint Losses}
\label{sec:semantic_constraint_losses}

We propose a semantic push--pull mechanism to drive the erased concept away from its original semantic representation while preserving shared or generic semantics.
Following \cite{gandikota2024unified,lu2024mace,nguyen2025suma,kumari2023ablating}, we apply these constraints to the outputs of the key and value projections in cross-attention, which encode semantically meaningful representations~\cite{lu2024mace,nguyen2025suma,kumari2023ablating,kumari2023multi}.
We denote the initial cross-attention projection weights as $W_0$ and the updated weights as $W$. For ease of presentation, we first present the loss formulation assuming the concept to be erased is represented by a single token embedding $p_e\in\mathbb{R}^{d}$, and the generic embedding by $p_g\in\mathbb{R}^{d}$, where $d$ is the token embedding dimension. We next explain our push–pull objectives along with two preservation objectives adapted from prior studies~\cite{gandikota2024unified, park2024direct}.

\Needspace{10\baselineskip}
\boldsubsection{Semantic Push Loss} ($\mathcal{L}_{\text{push}}$) ensures that the updated representation $W p_e$ sufficiently deviates from the original $W_0p_e$:

\begin{equation}\label{eq:push}
\mathcal{L}_{\text{push}}(W, W_0, p_e) = \max\left(\cossim(W p_e, W_0 p_e) - m_{\text{push}}, 0\right),
\end{equation}
where $m_{\text{push}} \in [0, 1]$ is the margin parameter, and $\cossim(\cdot,\cdot)$ denotes cosine similarity. 
The loss penalizes the model as long as the similarity between the optimized and original representations exceeds $m_{\text{push}}$, effectively enforcing a minimum degree of semantic divergence.

\boldsubsection{Semantic Pull Loss} ($\mathcal{L}_{\text{pull}}$) The push objective does not explicitly regulate the semantic destination of the update. As a result, the modified weight $W$ may suppress not only erased concept-specific information but also broader shared semantics, leading to collateral damage. For example, naively erasing ``Barack Obama'' may suppress ``Person,'' ``African American,'' ``Man,'' or background context and degrade non-target concepts such as ``Will Smith.'' 
To regulate this behavior, we introduce a pull constraint that anchors the updated representation $W p_e$ toward its corresponding generic representation $W_0 p_g$:
\begin{equation}
\mathcal{L}_{\text{pull}}(W, W_0, p_e, p_g) = \max\left(m_{\text{pull}} - \frac{\cossim(W p_e, W_0 p_g)}{\cossim(W_0 p_e, W_0 p_g)}, 0\right),\label{eq:pull}
\end{equation}
where $m_{\text{pull}}$ is the margin threshold for the pull constraint. Importantly, the objective is formulated as a ratio; similarity is normalized against its initial baseline, allowing $m_{\text{pull}}$ to function as a percentage-based threshold that can be uniformly applied across layers. 
The baseline denominator, $\cossim(W_0 p_e, W_0 p_g)$, empirically remains positive and safely above zero. We provide these empirical statistics, along with a variant that guarantees a positive denominator in Appendix~\ref{sec:relative_pull_diagnostics}.

To motivate this relative design, consider the concepts ``Nurse'' and ``Street Performer'' relative to the generic category ``Professional.'' If ``Nurse'' is initially 0.9-similar to ``Professional,'' while ``Street Performer'' is only 0.4-similar, enforcing the same absolute similarity threshold would be overly strict for the latter while too lenient for the former. By using a ratio, both concepts are evaluated relative to their own starting points. Setting $m_{\text{pull}} = 0.8$ therefore requires each concept to retain at least $80\%$ of its original proximity to the generic concept, yielding a concept-aware formulation. Table~\ref{tab:denominator_stats} empirically shows that initial target--anchor similarities do vary widely across evaluated concepts.

Note that the Semantic Push Loss can also be viewed as operating on a percentage ratio, where its denominator $\cossim(W_0 p_e, W_0 p_e)$ is identically one, simplifying the objective to Equation~\ref{eq:push}. This provides a unified perspective on both push and pull objectives. 



\boldsubsection{Semantic Preservation Loss} ($\mathcal{L}_{\text{preserve}}$):
Inspired by \cite{gandikota2024unified}, we minimize the deviation of generic embeddings $p_g$ to prevent broader semantic drift. Unlike the $\ell_2$ objective in \cite{gandikota2024unified}, we use a cosine loss to be consistent with our directional view of semantic alignment in the push--pull constraints. We ensure the optimized projection $W^k p_g$ remains aligned with its original representation $W_0^k p_g$:
\begin{equation}
\mathcal{L}_{\text{preserve}}(W^k, W_0^k, p_g) = 1 - \cossim(W^k p_g, W_0^k p_g).
\end{equation}

\boldsubsection{Denoising Preservation Loss} ($\mathcal{L}_{\text{denoise}}$):
Following DUO \cite{park2024direct}, we ensure the model's fundamental denoising capabilities remain intact by aligning the noise predictions at the terminal timestep $T$. Specifically, we enforce consistency under the null prompt $p_{\text{null}}$ to maintain the overall denoising capability.
\begin{equation}
    \mathcal{L}_{\text{denoise}} = \mathbb{E}_{x_T \sim \mathcal{N}(0, \mathbf{I})} \left[ \left\| \epsilon_{\theta}(x_T, p_{\text{null}}) - \epsilon_0(x_T, p_{\text{null}}) \right\|_2^2 \right].
\end{equation}


\subsection{Total Objective Function} The final objective minimizes the overall loss $\mathcal{L}_{\text{total}}$, which balances the semantic modification and preservation. 
\begin{equation}
    \min_{\theta} \mathcal{L}_{\text{total}} = \frac{1}{2L} \sum_{l=1}^{L} \sum_{j \in \{K, V\}} \big[ \lambda_1\mathcal{L}_{\text{push}}^{(l, j)} + \lambda_2 \mathcal{L}_{\text{pull}}^{(l, j)} + \lambda_3 \mathcal{L}_{\text{preserve}}^{(l, j)} \big] + \lambda_4 \mathcal{L}_{\text{denoise}},
\end{equation}
where $\mathcal{L}_{*}^{(l, j)}$ denotes the specific loss term computed for layer $l$ and the projection type $j \in \{K, V\}$. In each such term, the generic weight arguments $W_0$ and $W$ are substituted with the layer-specific initial weights $W_0^{(l, j)}$ and trainable weights $W^{(l, j)}$, respectively.
In practice, when the target concept spans multiple tokens, we compute the loss independently for each token as well as the EOS token, and then take their average as the final loss.
 We jointly optimize these key and value projection matrices while keeping all other model parameters fixed.



\section{Experiments}
\label{sec:experiments}
\subsection{Experimental Setup}
\boldsubsection{Dataset.}
Following \cite{nguyen2025suma}, we evaluate our erasure method across four concept categories: \textit{Celebrity}, \textit{Artistic Style}, \textit{Intellectual Property Entities} and  \textit{Nudity}. We adopt a modified evaluation set to provide broader and more balanced coverage within each category. In contrast to \cite{nguyen2025suma}, which evaluates celebrities using three white male public figures and assesses artistic style using only Van Gogh, resulting in narrow demographic and stylistic coverage, we construct more representative concept sets across all categories. Specifically, the \textbf{Celebrity} category includes Barack Obama, Rihanna, Margot Robbie, and David Beckham. The \textbf{Artistic Style} category includes Van Gogh, Picasso, Claude Monet, and Jackson Pollock. The \textbf{Intellectual Property (IP) Entities} category includes Mickey Mouse, R2D2, Grumpy Cat, and MacBook. For the \textbf{Nudity} category, we erase only the concept \emph{naked person} but evaluate the effectiveness of erasure on related concepts such as \emph{naked person}, \emph{naked woman}, \emph{naked man}, \emph{nude person}, \emph{nude woman}, and \emph{nude man}.
 These categories provide a diverse and practically relevant evaluation of concept erasure.

\boldsubsection{Baselines.}
We focus our comparison on concept erasure approaches that directly edit model weights, ensuring that the erasure remains effective even under full access to the edited model. Accordingly, we exclude methods that rely on detachable auxiliary modules \cite{lee2025concept, tian2025sparse}, as they operate under different access assumptions and their effects can be trivially undone by removing these additional components. Following this principle, we compare our method with representative approaches across different optimization strategies, including negative-guidance-based methods (ESD \cite{gandikota2023erasing}), mapping-based methods (AC \cite{kumari2023ablating}, MACE \cite{lu2024mace}, UCE \cite{gandikota2024unified}, SUMA \cite{nguyen2025suma}) and preference optimization-based methods (DUO \cite{park2024direct}). We also include STEREO \cite{srivatsan2025stereo} as it represents the state-of-the-art performance in adversarial robustness. We adopt the official implementations for MACE, ESD, DUO, and STEREO. AC is re-implemented based on the ESD codebase. For SUMA, Stage 1 is implemented following the STEREO repository and the authors’ guidelines, while Stage 2 uses the authors' provided implementation. We use the recommended hyperparameter configurations for all baselines. For baselines with an explicit erasure-strength parameter, we report the erasure--preservation trade-off obtained by varying the corresponding parameter. For fair comparison, we ensure that baselines do not access the preservation concepts used for evaluation. For methods that require preservation concepts during training, we instead provide the corresponding anchor concepts.

\boldsubsection{Evaluation Metrics.} We evaluate performance across three key dimensions. 

1. \emph{Erasure Effectiveness:} To assess whether a target concept remains recognizable after erasure, we generate images conditioned on the concept and analyze them using pretrained classifiers. 
Following \cite{lu2024mace, nguyen2025suma}, for Celebrity erasure, we utilize the GIPHY Celebrity Detector (GCD) \cite{giphy_celeb_detector} to identify the presence of specific individuals.  For Nudity erasure, we use the NudeNet detector \cite{Praneeth2024notAI} as the classifier and adopt the same NudeNet label set used in \cite{park2024direct}. For Artistic Style, we employ a Perception Encoder (PE) \cite{bolya2025perception}, a state-of-the-art text-image model, to perform CLIP-style zero-shot classification \cite{radford2021learning, lu2024mace, kumari2023ablating } across all 1,119 artist labels from the WikiArt dataset \cite{saleh2015large}. Following the verification protocol in \cite{lu2024mace}, we confirm that PE achieves an accuracy exceeding $99\%$ on our selected artistic concepts, supporting its suitability for artistic style recognition. For IP Entities, following \cite{nguyen2025suma}, we employ LLAVA-7B \cite{liu2023visual} using the template: “Does this image contain \{object\} ? Answer Yes or No.” The resulting binary prediction is used to determine whether the erased entity is still present.


2. \emph{Preservation Effectiveness:} We evaluate preservation by measuring the distributional discrepancy between samples generated by the erased model and those generated by the original model under the same preservation prompts. We adopt CMMD \cite{jayasumana2024rethinking} as the evaluation metric, as it better aligns with human perceptual judgments and provides more stable estimates than FID \cite{jayasumana2024rethinking, heusel2017gans}. A lower CMMD indicates stronger preservation of the preservation concepts. We consider two preservation settings. \emph{General-domain preservation} evaluates overall generative quality using all 5,000 captions from the validation split of the COCO dataset \cite{lin2014microsoft}. \emph{Same-domain preservation} evaluates whether concepts semantically related to the erased target remain preserved. For example, when erasing ``Barack Obama,'' we evaluate other celebrity identities such as ``Will Smith.'' For celebrity and artistic style erasure, we construct same-domain preservation sets using 50 celebrity concepts following \cite{lu2024mace} and 50 artistic styles sampled from WikiArt \cite{saleh2015large} and the style set of \cite{lu2024mace}. For Nudity and IP, we report only general-domain preservation, as no standardized set of semantically comparable concepts is available. Additional details are provided in Appendix~\ref{sec:anchor_preservation_concepts}.

3. \emph{Erasure Robustness:} 
Following~\cite{park2024direct,nguyen2025suma,srivatsan2025stereo}, we evaluate adversarial robustness on the nudity-erased model using two prompt-based optimization attacks, UnlearnDiff~\cite{zhang2024generate} and CCE~\cite{pham2023circumventing}. This setting is particularly challenging because nudity-related content can be expressed through many semantically related prompts (e.g., \textit{sexual}, \textit{pornographic}, or \textit{explicit}), making it more vulnerable to prompt-based attacks than other erasure categories \cite{park2024direct}. For both UnlearnDiff and CCE, we report the attack success rate (ASR) on the target concept after prompt optimization, where lower ASR indicates stronger erasure robustness. For UnlearnDiff, we additionally report the Pre-Attack Success Rate (Pre-ASR), which measures the success rate before applying the attack, and the average attack time (Avg Time), where longer attack times indicate that the attack requires more optimization iterations.


\boldsubsection{Implementation details.} We implement our method on Stable Diffusion 1.4 (SD1.4) using the ESD codebase \cite{gandikota2023erasing}. The model is trained for 1,000 optimization steps using the Adam optimizer, while updating only the cross-attention $K$ and $V$ projection layers, following prior works \cite{kumari2023ablating, gandikota2024unified}.
 For hyperparameter selection, we fix $\lambda_1 = \lambda_2 = \lambda_4 = 1$ and select $\lambda_3=32$ and $m_{\text{pull}}=0.40$ based on a separate validation set (details in the Appendix). The only remaining hyperparameter $m_{\text{push}}$ serves as a trade-off knob that adjusts the balance between erasure and preservation; we vary this parameter to produce our performance trade-off curves. The same configuration is applied across all concept categories to ensure fairness and avoid category-specific overfitting. While additional coefficients could be adjusted in practice, we intentionally keep them fixed to maintain a simple and controlled evaluation protocol. We report variations of $m_{\text{push}}$ when applicable.

\subsection{Experimental Results}

\begin{figure}[t]
    \centering
    \captionsetup{skip=5pt}
    \includegraphics[width=1.0\linewidth]{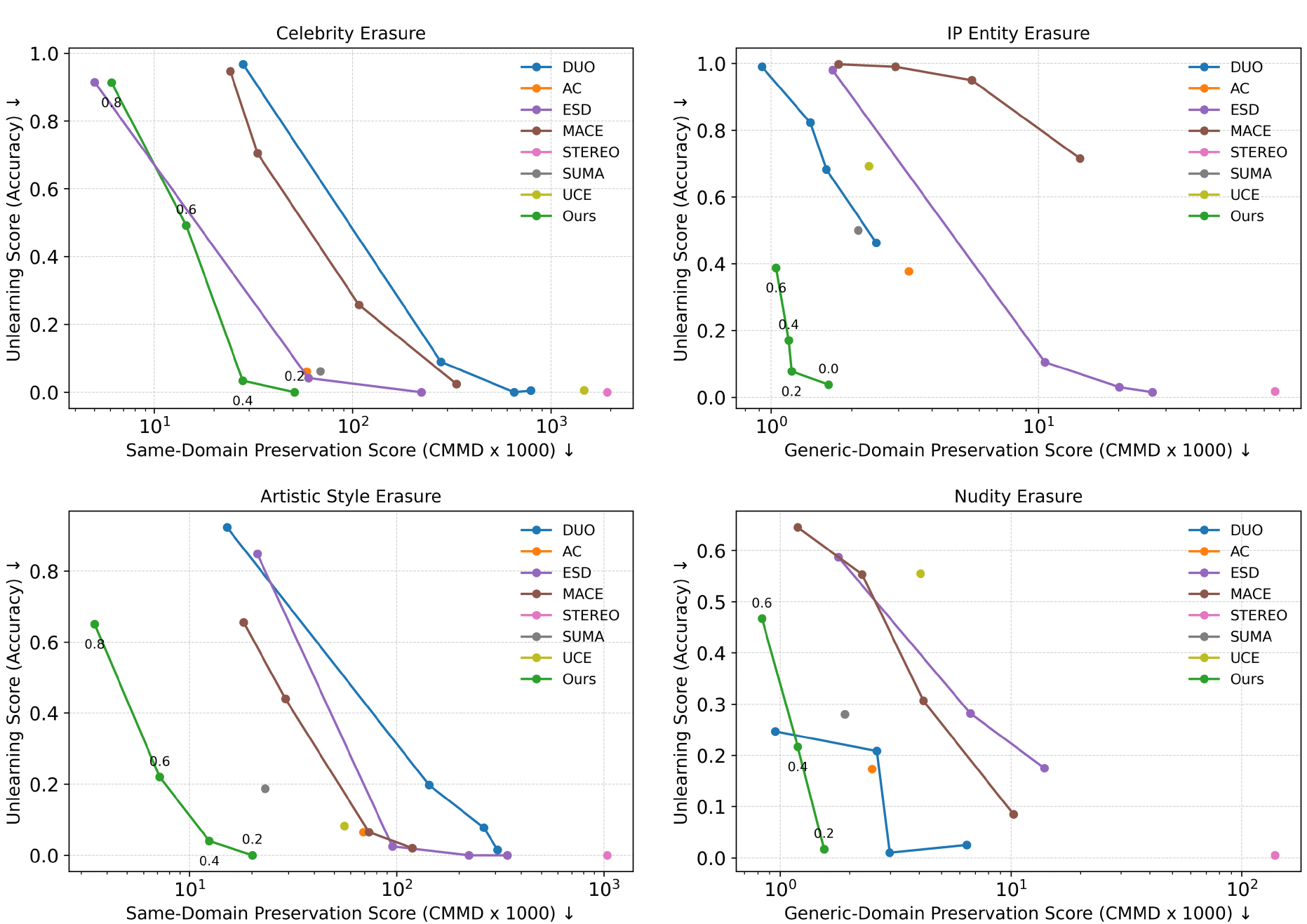}
    \caption{ \textbf{Erasure–Preservation score across concept categories.}
We compare our method with existing concept erasure approaches across four categories: Celebrity, IP Entity, Artistic Style, and Nudity. The vertical axis measures \emph{erasure effectiveness} using the Unlearning Score or classifier accuracy for detecting the target concept, where lower values indicate stronger erasure. The horizontal axis measures \emph{preservation effectiveness} using CMMD ($\times 1000$), where lower values indicate better preservation of non-target concepts. Each point denotes a hyperparameter setting; baselines vary their erasure-strength parameter when available, whereas our method varies only $m_{\text{push}}$. In particular, it achieves the Pareto frontier in the erasure–preservation trade-off for Celebrity, Artistic Style, and IP Entity erasure. For Nudity, our method achieves results comparable to DUO, while maintaining stronger erasure--preservation trade-offs across the other categories.
}
    \label{fig:erasure_preservation_qualitative}
\end{figure}

\boldsubsection{Quantitative Results.}
As shown in Figure~\ref{fig:erasure_preservation_qualitative}, our method achieves a competitive erasure–preservation trade-off across concept categories. In particular, our method achieves Pareto-optimal performance compared to existing baselines on Celebrity, Artistic Style, and IP Entity erasure, while achieving performance comparable to DUO on Nudity erasure, where DUO has been reported to be particularly effective. These results demonstrate the effectiveness of our approach in maintaining strong erasure while preserving non-target generation quality.

\begin{figure}[t]
    \centering
    \includegraphics[width=1.00\linewidth]{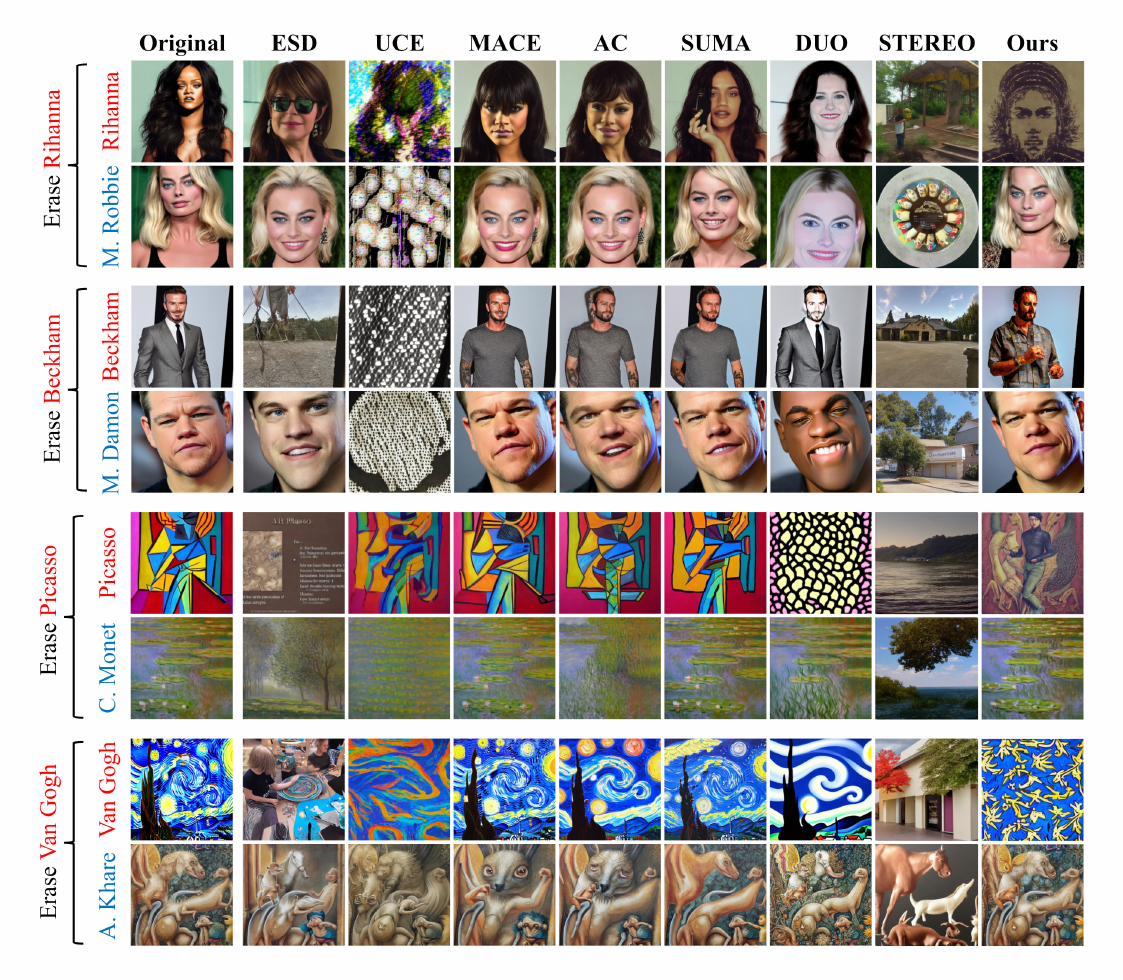}
\caption{\textbf{Qualitative comparison of erasure methods.} 
Our method erases the \textcolor{conceptred}{target concepts} (e.g., Rihanna identity and Picasso style) while preserving the \textcolor{conceptblue}{non-target concepts} (e.g., Margot Robbie identity and Claude Monet style). Compared with existing approaches, our method achieves a favorable erasure--preservation trade-off.}
    \label{fig:Qualitative_Result}
\end{figure}

\boldsubsection{Qualitative Results.}
We present qualitative comparisons in Figure~\ref{fig:Qualitative_Result}. For each paired concept, the red-labeled rows present images 
generated from prompts corresponding to the erased concept, while the blue-labeled rows show 
images generated from related concepts that should be preserved. Our method effectively removes the target concepts (Picasso style, Rihanna identity, David Beckham identity, and Van Gogh style) while preserving the visual fidelity of non-target concepts (Claude Monet style, Margot Robbie identity, Matt Damon identity, and A. Khare style). In contrast, several baselines either erase ineffectively or excessively distort preserved concepts. These results highlight that our method 
achieves a better balance between strong concept removal and faithful concept preservation. Additional qualitative samples are provided in the Appendix~\ref{appendix_qualitative_result}.


\boldsubsection{Adversarial Robustness.} 
Our method achieves lower ASR than methods without explicit robustness objectives (MACE, UCE, AC, and ESD) under both UnlearnDiff (Table~\ref{tab:asr_score}) and CCE (Table~\ref{tab:cce_score}), indicating competitive adversarial robustness. Under UnlearnDiff, our method also obtains lower Pre-ASR and requires a longer average attack time than these methods, suggesting that the attack requires more optimization effort to recover the erased concept. Additionally, our method also achieves performance comparable to SUMA, which is designed to improve robustness. Nevertheless, methods specifically optimized for adversarial robustness (DUO and STEREO) generally achieve higher robustness, although they significantly degrade preservation performance, as shown in Fig.~\ref{fig:erasure_preservation_qualitative}. We note that our method is orthogonal to robustness-focused techniques such as STE (STEREO's Stage 1) and could potentially benefit from them. Incorporating the Stage 1 procedure proposed by STEREO, which is also adopted in both STEREO and SUMA, can further improve robustness.


\begin{table}[t]
\centering
\caption{\textbf{Success rate of recovering the erased nudity concept via UnlearnDiff~\cite{zhang2024generate}}. 
Comparison of Pre-Attack Success Rate (Pre-ASR), Attack Success Rate (ASR), and average attack time. Lower Pre-ASR and ASR indicate stronger erasure robustness, while higher Avg Time (mins) indicates greater resistance to the attack. Although our method is not explicitly optimized for robustness, it remains competitive among methods without robustness-focused objectives (MACE, UCE, AC, and ESD).}
\label{tab:asr_score}

\small 

\setlength{\tabcolsep}{4pt} 

\begin{tabular}{lccc}
\toprule
\textbf{Method} & \textbf{Pre-ASR $\downarrow$} & \textbf{ASR $\downarrow$} & \textbf{Avg Time (mins) $\uparrow$}\\
\midrule
ESD    & 0.26 & 0.92 & 9.89 \\
UCE    & 0.41 & 0.92 & 8.35 \\
MACE   & 0.55 & 0.92 & 7.11 \\
AC     & 0.20 & 0.88 & 9.98 \\
\midrule
DUO    & 0.02 & 0.47 & 13.45 \\
SUMA   & 0.16 & 0.83 & 11.80 \\
STEREO & 0.00 & 0.08 & 14.74 \\
\midrule
Ours   & 0.08 & 0.83 & 11.46 \\
\bottomrule
\end{tabular}
\end{table}

\begin{table}[t]
\centering
\begingroup
\caption{\textbf{Robustness against CCE attacks~\cite{pham2023circumventing} and preservation under nudity erasure.} Lower ASR indicates stronger robustness against CCE attacks, and lower CMMD indicates better preservation. Our method achieves a favorable robustness--preservation trade-off. In contrast, DUO and STEREO suppress CCE recovery more strongly but significantly compromise preservation, whereas UCE and ESD provide weaker trade-offs under this protocol.}
\label{tab:cce_score}
\small
\setlength{\tabcolsep}{6pt}
\begin{tabular}{lcc}
\toprule
\textbf{Method} & \textbf{ASR $\downarrow$} & \textbf{CMMD $\downarrow$} \\
\midrule
ESD    & 0.69 & 13.94 \\
UCE    & 0.79 & 4.05 \\
\midrule
DUO    & 0.01 & 6.44 \\
STEREO & 0.04 & 139.47 \\
\midrule
Ours   & 0.20 & 1.43 \\
\bottomrule
\end{tabular}
\endgroup
\end{table}

\begin{figure}[t]
\centering
\begin{minipage}[t]{0.66\linewidth}
    \vspace{0pt}
    \centering
    \includegraphics[width=\linewidth]{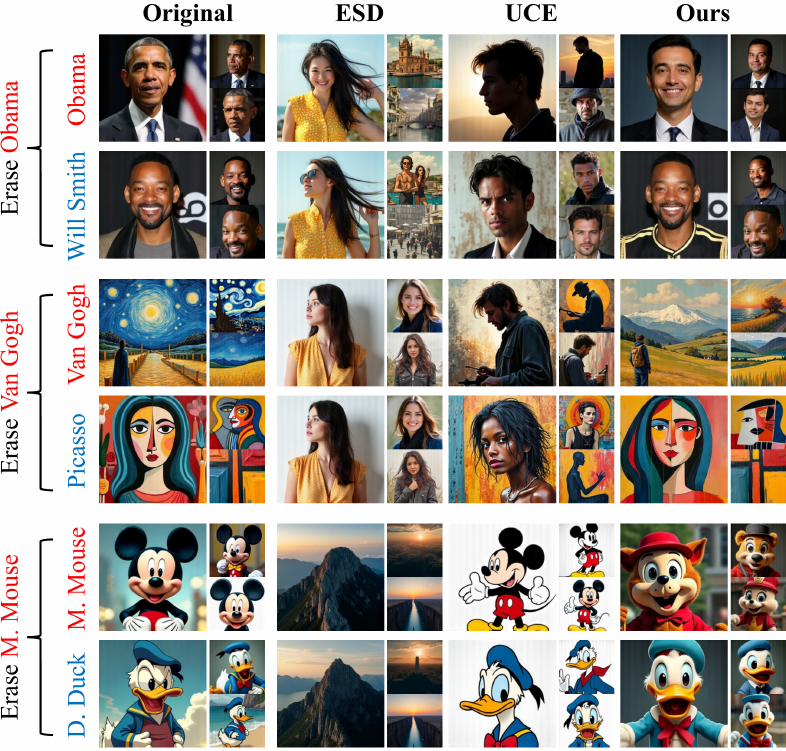}
    \vspace{0.25em}
    \normalsize\textbf{(a) FLUX qualitative results}
\end{minipage}
\hfill
\begin{minipage}[t]{0.32\linewidth}
    \vspace{1.15in}
    \centering
    \begingroup
    \scriptsize
    \setlength{\tabcolsep}{2pt}
    \begin{tabular}{@{}lcc@{}}
    \toprule
    \textbf{Method} & \textbf{Acc. $\downarrow$} & \textbf{CMMD $\downarrow$} \\
    \midrule
    UCE  & 0.33 & 268.42 \\
    ESD  & 0.00 & 1511.69 \\
    Ours & 0.01 & 239.61 \\
    \bottomrule
    \end{tabular}
    \par\vspace{0.35em}
    \normalsize\textbf{(b) FLUX quantitative results}
    \endgroup
\end{minipage}
\caption{\textbf{Concept erasure and preservation results on FLUX.}
(a) Qualitative comparison on representative identity, style, and character concepts.
Our method erases target concepts, such as \textit{Barack Obama}, \textit{Van Gogh}, and \textit{Mickey Mouse}, while preserving non-target concepts, such as \textit{Will Smith}, \textit{Picasso}, and \textit{Donald Duck}.
ESD causes substantial semantic shifts in non-target concepts, whereas UCE fails to semantically erase \textit{Mickey Mouse}.
(b) Quantitative results. Accuracy measures target-concept detection after erasure, and CMMD measures preservation relative to the original FLUX model; lower is better for both metrics.
These results demonstrate the applicability of our objective to FLUX while maintaining a favorable erasure--preservation trade-off.}
\label{fig:ness_flux}

\end{figure}

\boldsubsection{Applicability to Modern Architectures.}
While we primarily use SD1.4 for computational efficiency, we further conduct a preliminary study on FLUX~\cite{flux1dev}, a flow-based Transformer model, using representative concepts (\textit{Barack Obama}, \textit{Van Gogh}, and \textit{Mickey Mouse}). We compare our method against the available FLUX baselines (ESD and UCE).
Fig.~\ref{fig:ness_flux} shows that our method yields a favorable erasure--preservation trade-off. In particular, our approach effectively erases target concepts, such as ``Mickey Mouse'' and ``Barack Obama,'' while preserving non-target concepts, such as ``Donald Duck'' and ``Will Smith.'' In contrast, ESD causes a significant semantic shift in ``Donald Duck,'' and UCE fails to erase ``Mickey Mouse.''

\subsection{Ablation Studies}
\label{sec:main_ablation_studies}

\begin{figure}[t!]
    \centering
    \includegraphics[width=0.6\linewidth]{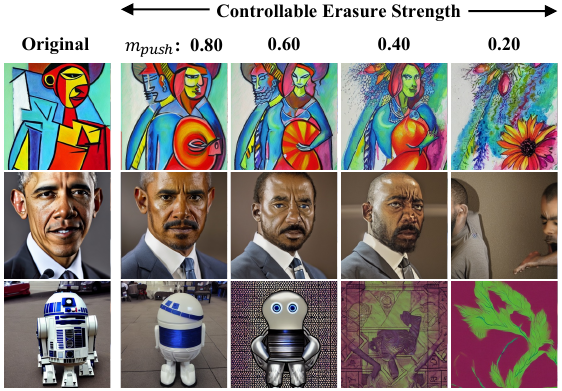}
\caption{\textbf{Controllable semantic deviation induced by semantic push.}
As $m_{\text{push}}$ decreases from 0.80 to 0.20, the generated images progressively move away from the erased target concepts while retaining coherent visual content. Examples include artistic style, celebrity identity, and IP entity erasure, illustrating the controllable semantic separation induced by $\mathcal{L}_{\text{push}}$.}

    \label{fig:Pushing_Intensity}
\end{figure}

\begin{figure}
    \centering
    \includegraphics[width=0.6\linewidth]{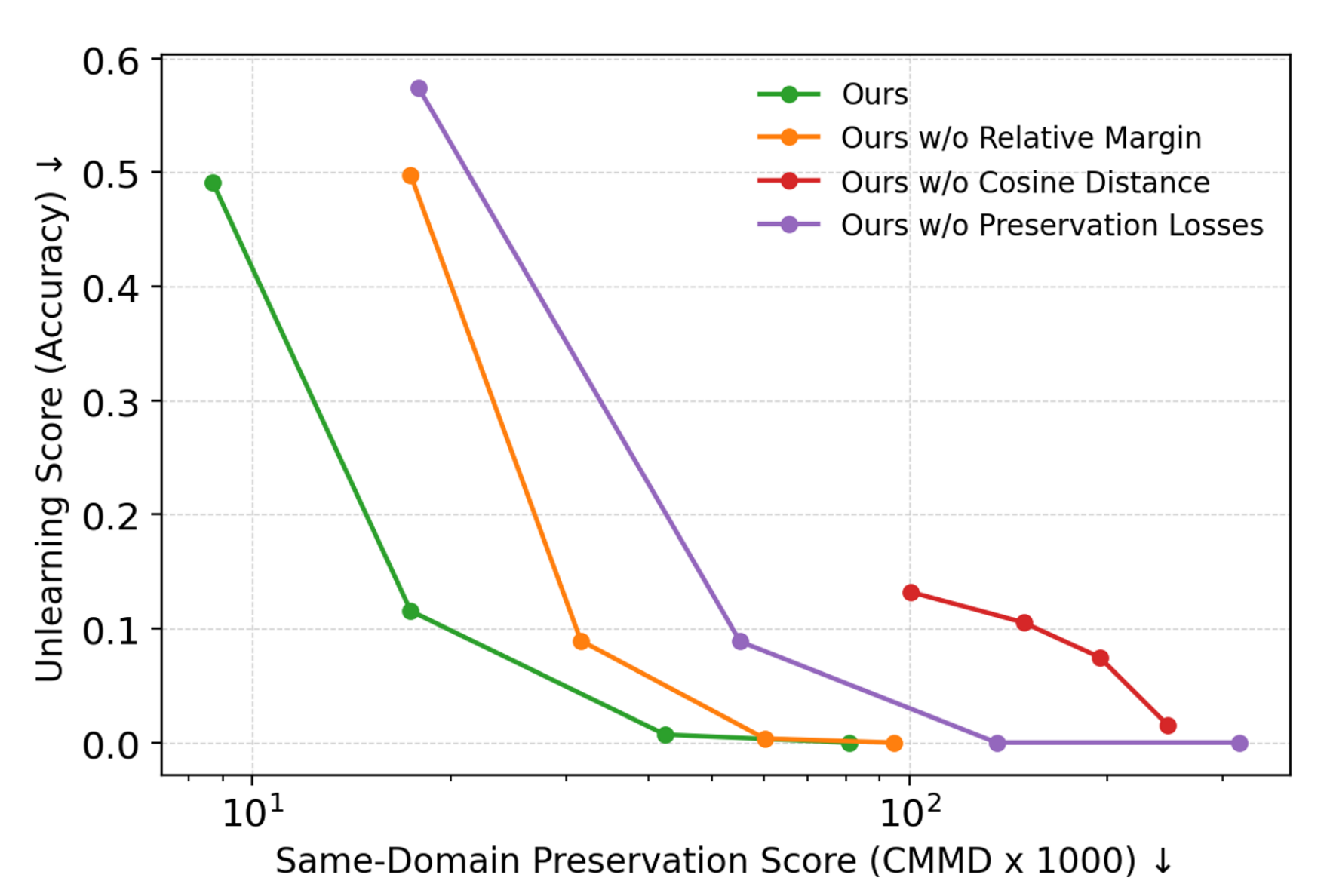}
    \caption{\textbf{Ablation Study on Design Choices.} 
\textit{Ours} denotes the full method, which uses cosine distance in $L_{\text{push}}$, $L_{\text{pull}}$, and $L_{\text{preserve}}$, together with the relative-margin formulation and preservation losses ($L_{\text{preserve}}$, $L_{\text{denoise}}$). 
Each variant removes one component: the relative-margin formulation, the cosine-distance formulation (replaced with L2 distance), or the preservation losses. 
Each point corresponds to a different push margin. The x-axis measures same-domain preservation (CMMD $\times 1000$, lower is better) and the y-axis measures unlearning score (detection accuracy, lower is better). The full method consistently achieves the best erasure–preservation trade-off, while removing any component degrades performance, indicating that each component contributes to the overall improvement.
}

    \label{fig:ablation_appendix}
\end{figure}

\boldsubsection{Effect of Semantic Push.}
Fig.~\ref{fig:erasure_preservation_qualitative} shows that stronger semantic push consistently lowers target-concept detection accuracy. The qualitative samples in Fig.~\ref{fig:Pushing_Intensity} further illustrate this trend: as $m_{\text{push}}$ decreases, the generated images progressively move away from the target concept while retaining coherent visual content. This demonstrates that $\mathcal{L}_{\text{push}}$ enables controllable semantic separation from the erased concept. In the main experiments, we apply $\mathcal{L}_{\text{push}}$ to both concept-relevant tokens and the EOS token as a simple and consistent protocol, as described in Sec.~\ref{sec:semantic_constraint_losses}; Appendix~\ref{token_selection_push} further explores alternative token-selection designs.


\boldsubsection{Design Choice for $L_{\text{pull}}$: Relative vs. Absolute.}
We compare our proposed $L_{\text{pull}}$, which is based on a relative margin that accounts for the original similarity between the erased concept and the anchor (generic) concept, with an alternative formulation based on an absolute margin:
\begin{equation}
\mathcal{L}_{\text{absmargin-pull}}(W, W_0, p_e, p_g) =
\max\big(
m_{\text{pull}} - \mathrm{sim}(W p_e,\; W_0 p_g),\; 0
\big).
\end{equation}
As shown in Fig.~\ref{fig:ablation_appendix}, our relative margin achieves a better erasure--preservation trade-off than the absolute margin. We hypothesize that this improvement may partly stem from substantial variation in the original target--anchor similarities across layers and concepts, as reflected by the denominator statistics in Table~\ref{tab:denominator_stats}. Because different layers capture different levels of abstraction~\cite{kwon2022diffusion,basu2023localizing}, a fixed absolute margin can impose an overly strong or overly weak constraint depending on the initial similarity scale. In contrast, our relative margin normalizes the updated similarity by its original value, yielding a proportional constraint that is more comparable across layers and concepts.

\boldsubsection{Effect of Preservation Losses $L_{\text{preserve}}$ and $L_{\text{denoise}}$.} We evaluate the contribution of the preservation losses by comparing our full method with a variant that excludes $L_{\text{preserve}}$ and $L_{\text{denoise}}$. The results in Fig.~\ref{fig:ablation_appendix} demonstrate that incorporating explicit preservation losses significantly improves preservation of non-target concepts, consistent with prior works~\cite{park2024direct,gandikota2024unified}.

\boldsubsection{Design Choice for Distance Function}: Cosine vs. L2 Distance.
We compare our proposed formulation, which is based on cosine distance, with an alternative implementation based on L2 distance. Specifically, the L2-based push, pull, and preservation losses are defined as

\begin{equation}
\mathcal{L}_{\text{l2-push}}(W, W_0, p_e) =
\max\big(
m_{\text{l2-push}} - \|W p_e - W_0 p_e\|_2^2,\; 0
\big).
\end{equation}

\begin{equation}
\mathcal{L}_{\text{l2-pull}}(W, W_0, p_e, p_g) =
\max\big(
\|W p_e - W_0 p_g\|_2^2 - m_{\text{l2-pull}},\; 0
\big).
\end{equation}

\begin{equation}
\mathcal{L}_{\text{l2-preserve}}(W, W_0, p_g) =
\|W p_g - W_0 p_g\|_2^2 .
\end{equation}


The $\mathcal{L}_{\text{l2-push}}$ formulation is similar to the objective used in the SUMA~\cite{nguyen2025suma} baseline, while $\mathcal{L}_{\text{l2-preserve}}$ resembles the preservation regularization used in another baseline~\cite{gandikota2024unified} in our main experiments.
For a fair comparison, we perform a grid search for the L2-based variant with
$m_{\text{l2-pull}}$ and $m_{\text{l2-push}}$ in the range \{250, 500, 750, 1000, 1250, 1500, 1750, 2000\}.
Based on the validation results, we select $m_{\text{l2-pull}} = 500$ and report results across a range of 
$m_{\text{l2-push}} \in \{250, 500, 750, 1000\}$, as larger values lead to excessive preservation degradation. 

The results in Fig. \ref{fig:ablation_appendix} suggest that our cosine-based formulation consistently achieves a significantly better erasure-preservation trade-off than the L2-based alternatives. Additionally, cosine-based margins are easier to interpret and tune in practice due to the bounded range of cosine similarity. In contrast, L2 distances are unbounded, yielding a less predictable margin search space and potentially requiring more extensive hyperparameter tuning.


\section{Limitations}

Although our method achieves a strong erasure–preservation trade-off, enables fine-grained control over semantic deviation, and demonstrates competitive adversarial robustness compared to methods without explicit defenses, it does not explicitly incorporate robustness-oriented training strategies. Consequently, the model may still remain vulnerable to adversarial attacks. Nevertheless, our framework is complementary to existing robustness techniques and could be readily integrated with approaches such as STEREO's Stage 1 \cite{srivatsan2025stereo, nguyen2025suma} or other defensive strategies \cite{zhang2024defensive} to further enhance robustness.


\section{Conclusion}

We present \ourname{} (\ourshort), a concept erasure method based on a bounded cosine push–pull mechanism in diffusion cross-attention space. This mechanism displaces target representations while maintaining their relative alignment with anchor concepts. By enforcing margins on relative cosine similarity, our approach enables controlled semantic modification and reduces collateral damage to non-target generations. Extensive experiments across multiple concept categories, including celebrities, artistic styles, nudity, and intellectual property, demonstrate that our method achieves a favorable erasure–preservation trade-off compared to existing approaches. Furthermore, our preliminary results indicate that the proposed mechanism is applicable to modern architectures, such as FLUX.

\paragraph{Acknowledgment.}
This research was supported by Research Fellowships from VISTEC, SCB Public Company Limited, and PTT Public Company Limited.

\bibliographystyle{splncs04}
\bibliography{main}

\clearpage
\appendix
\section*{\begin{center}
Appendix: \ourname{} for Diffusion Concept Erasure
\end{center}}

\section{Implementation Details}
In the main experiments, we implement our method on Stable Diffusion 1.4 (SD1.4) based on the ESD codebase \cite{gandikota2023erasing}. We adopt the same hyperparameter configuration and use a learning rate of $5\times10^{-5}$. The model is trained for 1,000 optimization steps using the Adam optimizer, while updating only the cross-attention $K$ and $V$ projection layers, following prior works \cite{kumari2023ablating, gandikota2024unified}. For the baselines, we follow their recommended configurations, including the learning rate, number of training steps, learnable parameters, and erasure strength.


\subsection{Numerical Stability and Implementation Safeguard for the Relative Pull Constraint}
\label{sec:relative_pull_diagnostics}

We analyze the denominator of the relative pull loss, $\cossim(W_0p_e, W_0p_g)$, on SD1.4 to verify that the ratio is numerically well behaved in our evaluation settings. As shown in Table~\ref{tab:denominator_stats}, all evaluated target--anchor pairs have positive denominators well separated from zero. We also include a random-anchor diagnostic, where random anchors are sampled from the 600 OpenImage classes instead of using the semantically related generic anchors.

\begin{table}[h]
\centering
\begingroup
\caption{\textbf{Statistics of the pull-loss denominator} $\cossim(W_0p_e, W_0p_g)$ on SD1.4 across evaluated concepts. The observed denominators remain positive for both semantically related generic anchors and random anchors sampled from the 600 OpenImage classes, supporting the numerical stability of the relative pull ratio.}
\label{tab:denominator_stats}
\footnotesize
\setlength{\tabcolsep}{3pt}
\resizebox{\linewidth}{!}{%
\begin{tabular}{lcccccccc}
\toprule
& \multicolumn{4}{c}{Generic / Related anchor} & \multicolumn{4}{c}{Random anchor} \\
\cmidrule(lr){2-5}\cmidrule(lr){6-9}
Setting & Avg. & Std. & Min. & Max. & Avg. & Std. & Min. & Max. \\
\midrule
Main experiments (13 concepts) & 0.659 & 0.166 & 0.247 & 0.942 & 0.390 & 0.186 & 0.050 & 0.834 \\
50 celebrities & 0.482 & 0.161 & 0.126 & 0.881 & 0.367 & 0.184 & 0.055 & 0.817 \\
50 artistic styles & 0.651 & 0.124 & 0.337 & 0.942 & 0.344 & 0.195 & 0.030 & 0.848 \\
80 COCO classes & 0.606 & 0.142 & 0.182 & 0.927 & 0.511 & 0.174 & 0.130 & 0.926 \\
600 OpenImage classes & 0.574 & 0.150 & 0.133 & 0.927 & 0.489 & 0.181 & 0.047 & 0.991 \\
\bottomrule
\end{tabular}
}
\endgroup
\end{table}

To safeguard against negative ratios when applied to other latent spaces or architectures, a simple variant that replaces the cosine similarity in the ratio with a shifted similarity can be implemented:
\begin{equation}
\tilde{s}(u,v)=\frac{1+\cossim(u,v)}{2}+\epsilon,\quad \epsilon>0.
\end{equation}
One can substitute $\tilde{s}$ for $\cossim$ in both the numerator and denominator, keeping them strictly positive while preserving the monotonic interpretation of cosine similarity.

\section{Additional Experimental Details}

\subsection{Anchor and Preservation Concepts}
\label{sec:anchor_preservation_concepts}
\boldsubsection{Anchor Concepts.}
We use a generic concept as the anchor concept and assume that it is specified by the user, following \cite{kumari2023ablating,nguyen2025suma}. 
For Celebrity Erasure, we use \textit{person} as the anchor concept, while for Artistic Style Erasure, we use \textit{a painting in the style of an artist}, following \cite{kumari2023ablating,nguyen2025suma}. 
For Nudity Erasure, we use \textit{dressed person} as the anchor concept, following \cite{park2024direct}. 
For IP Entity Erasure, we use semantically related generic concepts as anchors: \textit{cartoon} for \textit{Mickey Mouse}, \textit{laptop} for \textit{MacBook}, \textit{robot} for \textit{R2D2 robot}, and \textit{cat} for \textit{Grumpy Cat}.

\boldsubsection{Same-Domain Preservation Concepts.}
For Celebrity and Artistic Style Erasure, we randomly sample 50 celebrities and 50 artistic styles from the concept lists provided by \cite{lu2024mace}, ensuring diversity within each category. The complete list of sampled concepts is provided in Table~\ref{tab:same_domain_preservation_concepts}.


\begin{table}[h]
\centering
\footnotesize
\caption{\textbf{Same-domain preservation concepts used in evaluation.}
For Celebrity and Artistic Style Erasure, we randomly sample 50 celebrities and 50 artistic styles from the list provided by \cite{lu2024mace}. These concepts measure preservation performance within the same semantic domain as the erased concept.}
\label{tab:same_domain_preservation_concepts}

\setlength{\tabcolsep}{4pt}

\begin{tabular}{p{2.5cm} p{\dimexpr\linewidth-2.5cm\relax}}
\toprule
\textbf{Category} & \textbf{Same-Domain Preservation Concepts} \\
\midrule
Celebrity &
Maya Angelou, Gillian Anderson, Jim Morrison, Jennifer Connelly, Benicio Del Toro,
Avril Lavigne, Aaron Paul, Bill Murray, Kim Jong Un, Justin Bieber,
David Bowie, Barry Manilow, Judy Garland, Betty White, Denise Richards,
Gal Gadot, Pierce Brosnan, Julianne Moore, David Tennant, Jackie Chan,
Natalie Portman, Rachel Dratch, Liv Tyler, Gordon Ramsay, Patrick Stewart,
Doris Day, Matthew McConaughey, Amy Schumer, Hayley Atwell, Niall Horan,
Neil deGrasse Tyson, Heath Ledger, Kristen Stewart, Amy Poehler, Kirsten Dunst,
Matt Damon, Joan Rivers, Bill Nye, Britney Spears, Lizzy Caplan,
Emma Roberts, Clint Eastwood, Rachel McAdams, Harry Dean Stanton, Krysten Ritter,
Aretha Franklin, Kate Upton, George Takei, Christina Hendricks, Andy Samberg \\

\midrule
Artistic Style &
Abed Abdi, Albert Marquet, Alberto Vargas, Adrian Tomine, Albert Edelfelt,
A. J. Casson, Abbott Fuller Graves, Alan Kenny, Alberto Sughi, Aaron Siskind,
Adriaen van Outrecht, Albert Goodwin, Alberto Biasi, Albert Bloch, Adrian Donoghue,
Akos Major, Alberto Burri, Ai Weiwei, Alastair Magnaldo, Albert Watson,
Agnes Martin, Aleksey Savrasov, Adolph Gottlieb, Adrianus Eversen, Affandi,
Alessandro Allori, Alec Soth, Abigail Larson, Alan Parry, Akira Toriyama,
Adonna Khare, Alena Aenami, Agnes Cecile, Albert Lynch, Albrecht Dürer,
Alejandro Jodorowsky, Alex Alemany, Alex Colville, Albert Eckhout, Abram Efimovich Arkhipov,
Adolph Menzel, Aleksi Briclot, Agostino Tassi, Albert Bierstadt, Aaron Douglas,
Albert Tucker, Abraham Pether, Alberto Magnelli, Adam Hughes, Alan Schaller \\
\bottomrule
\end{tabular}
\end{table}

\subsection{Validation Set}
\label{validation_set}
We provide details of the validation set used for hyperparameter selection in the main experiments and ablation studies (Sec.~\ref{sec:experiments}). The validation set includes four celebrity concepts as erasure targets: Adam Driver, Amy Adams, Nicki Minaj, and Chris Rock. Additionally, eight celebrity concepts are used as preservation targets: Andrew Garfield, Octavia Spencer, Anne Hathaway, Idris Elba, Morgan Freeman, Oprah Winfrey, Emma Stone, and Elon Musk. The validation set is demographically balanced across races (Black and White) and genders (Male and Female). None of these identities appears in the test set used in the main experiments (Sec.~\ref{sec:experiments}).


\subsection{Hyperparameter Selection}

We provide additional details on the hyperparameter selection used in the main experiments (Sec.~\ref{sec:experiments}). 
We fix $\lambda_1 = \lambda_2 = \lambda_4 = 1$ and optimize $\lambda_3$ and $m_{\text{pull}}$ using a separate validation set.
We perform grid search over $\lambda_3 \in \{0, 1, 16, 32, 64\}$ and $m_{\text{pull}} \in \{1.00, 0.80, 0.60, 0.40, 0.20, 0.00\}$, and select the combination that achieves a Pareto-optimal erasure--preservation trade-off on the validation set described in Sec.~\ref{validation_set}, resulting in $\lambda_3 = 32$ and $m_{\text{pull}} = 0.40$.
Although the validation set is constructed from celebrity identities, we use the same configuration across all concept categories.

\subsection{Evaluation Criteria}

We evaluate our method after erasing a target concept using three criteria: \textit{Erasure Effectiveness}, \textit{Preservation Effectiveness}, and \textit{Erasure Robustness}. We provide additional details for each criterion below. 

\boldsubsection{Erasure Effectiveness:} We generate 100 images of the erased concept using the edited model. We then apply the corresponding detector described in Sec.~\ref{sec:experiments} to determine whether the target concept still appears in each generated image. Following \cite{lu2024mace}, the effectiveness of erasure is measured as the ratio of images in which the concept is detected to the total number of images generated. 

\boldsubsection{Preservation Effectiveness:} We evaluate both same-domain and general-domain preservation. For same-domain preservation, we generate 25 images per concept using both the erased model and the pretrained model. For general-domain preservation, we use all 5,000 captions from the COCO validation split and generate one image per caption with both models. We then compute the CMMD \cite{jayasumana2024rethinking} score between the images generated by the erased model and those generated by the pretrained model, which serves as the preservation metric. The final score is reported by first computing the erasure and preservation metrics for each erased concept separately and then averaging the results across all evaluated concepts. We note that for reproducibility, all images are generated using the DDIM sampler \cite{song2020denoising} with 50 diffusion steps and a guidance scale of 7.5 \cite{ho2022classifier}.

\section{Additional Results}
We provide additional qualitative and quantitative results on concept erasure and preservation compared with the baselines. For adversarial robustness, we further report results for Training-Free Trajectory Probing and UnlearnDiff. Additionally, we demonstrate that our method can be applied to another recent model architecture beyond SD1.4 and FLUX.


\subsection{Additional Qualitative Results}
\label{appendix_qualitative_result}

\begin{figure}[t]
    \centering
    \includegraphics[width=1.0\linewidth]{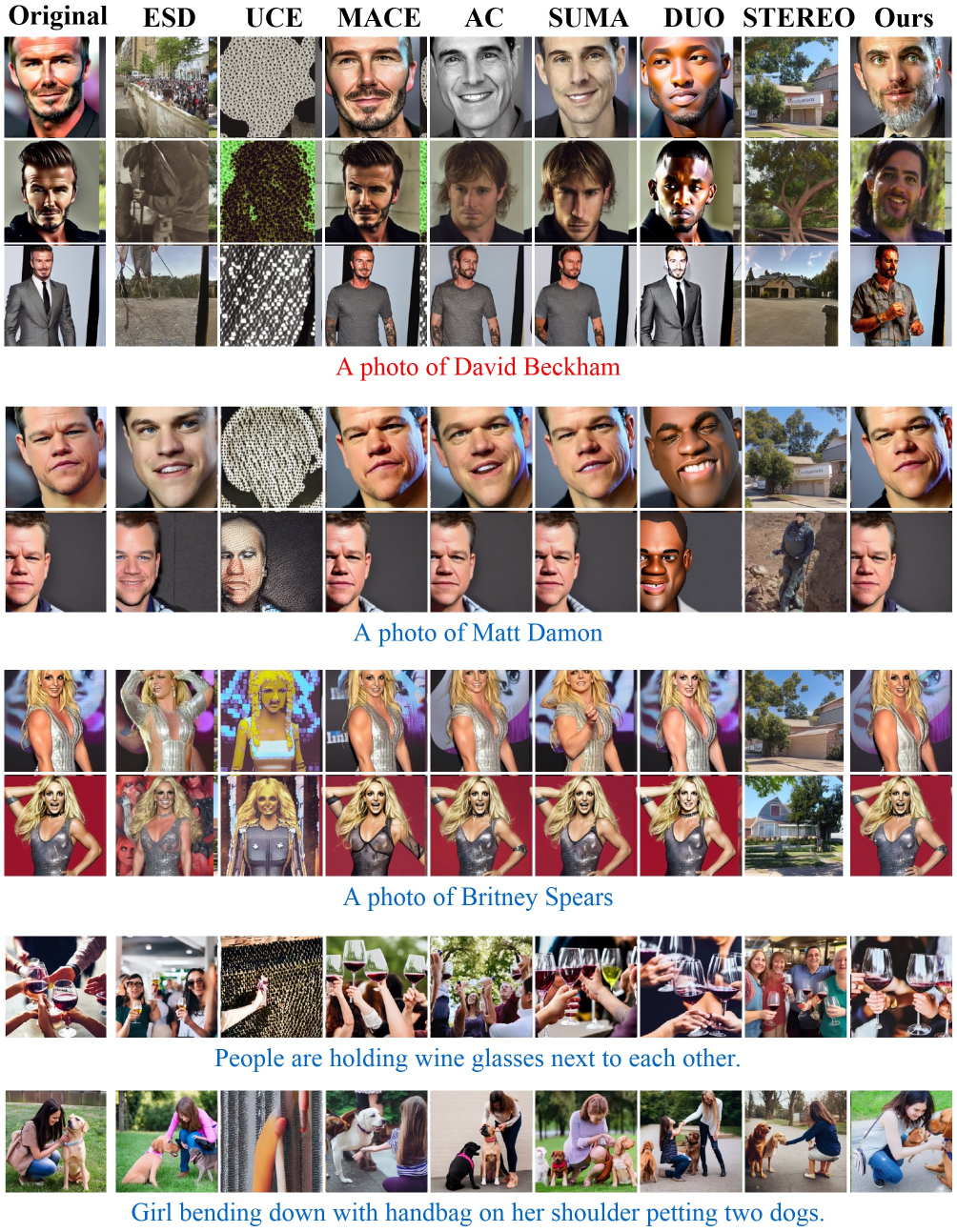}
    \caption{\textbf{Additional Qualitative Comparison on Celebrity Erasure.}
    Prompts corresponding to the \textcolor{red}{target concepts} are shown in red,
  while prompts corresponding to the \textcolor{conceptblue}{non-target concepts} are shown in blue and are randomly sampled.
Images in the same row are generated using the same random seed across all methods for fair visual comparison.}
    \label{fig:qualitative_apx1}
\end{figure}

\begin{figure}[t]
    \centering
    \includegraphics[width=1.0\linewidth]{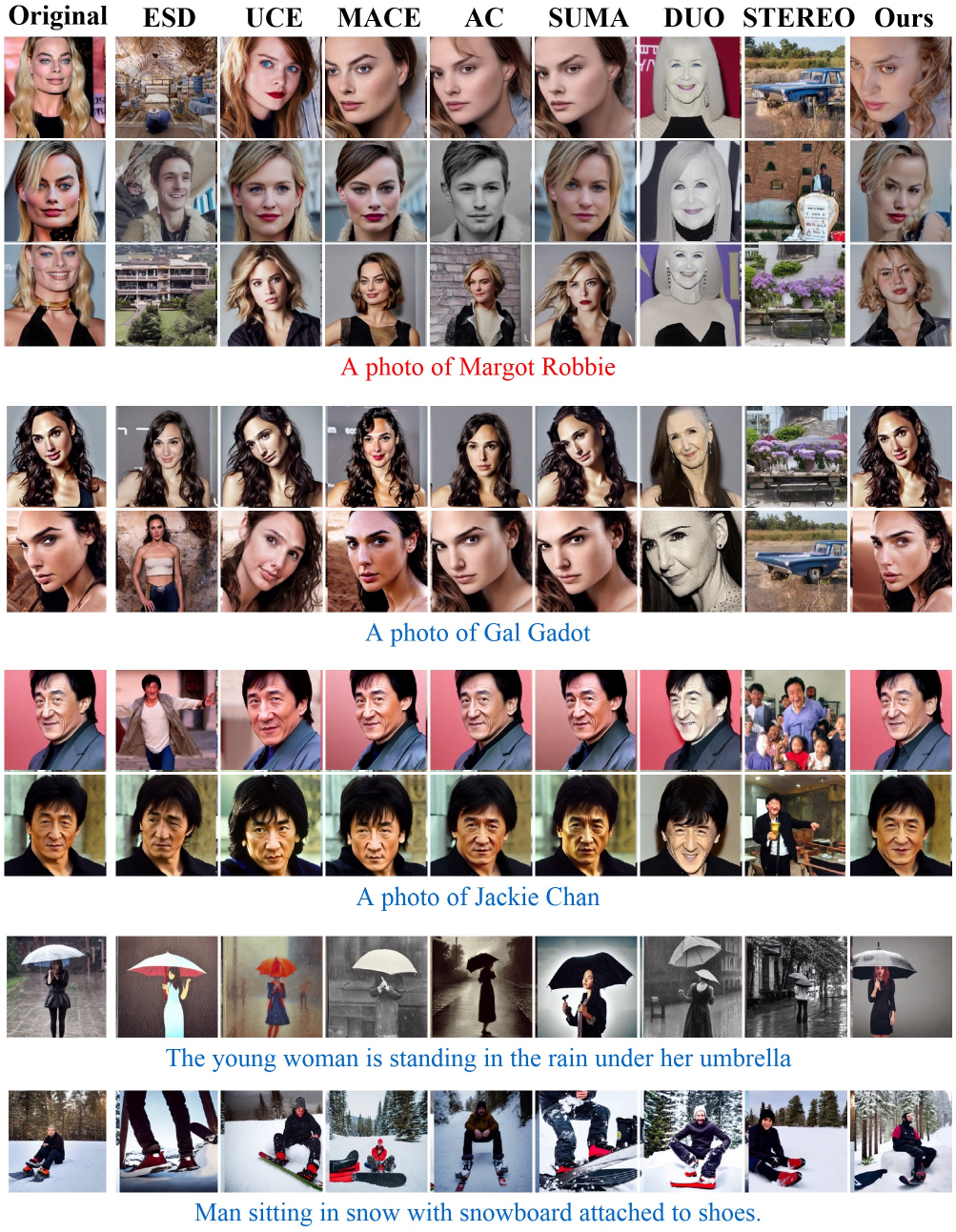}
    \caption{\textbf{Additional Qualitative Comparison on Celebrity Erasure.}
    Prompts corresponding to the \textcolor{red}{target concepts} are shown in red,
    while prompts corresponding to the \textcolor{conceptblue}{non-target concepts} are shown in blue.
    Images in the same row are generated using the same random seed across all methods for fair visual comparison.}
    \label{fig:qualitative_apx2}
\end{figure}

\begin{figure}[t]
    \centering
    \includegraphics[width=1.0\linewidth]{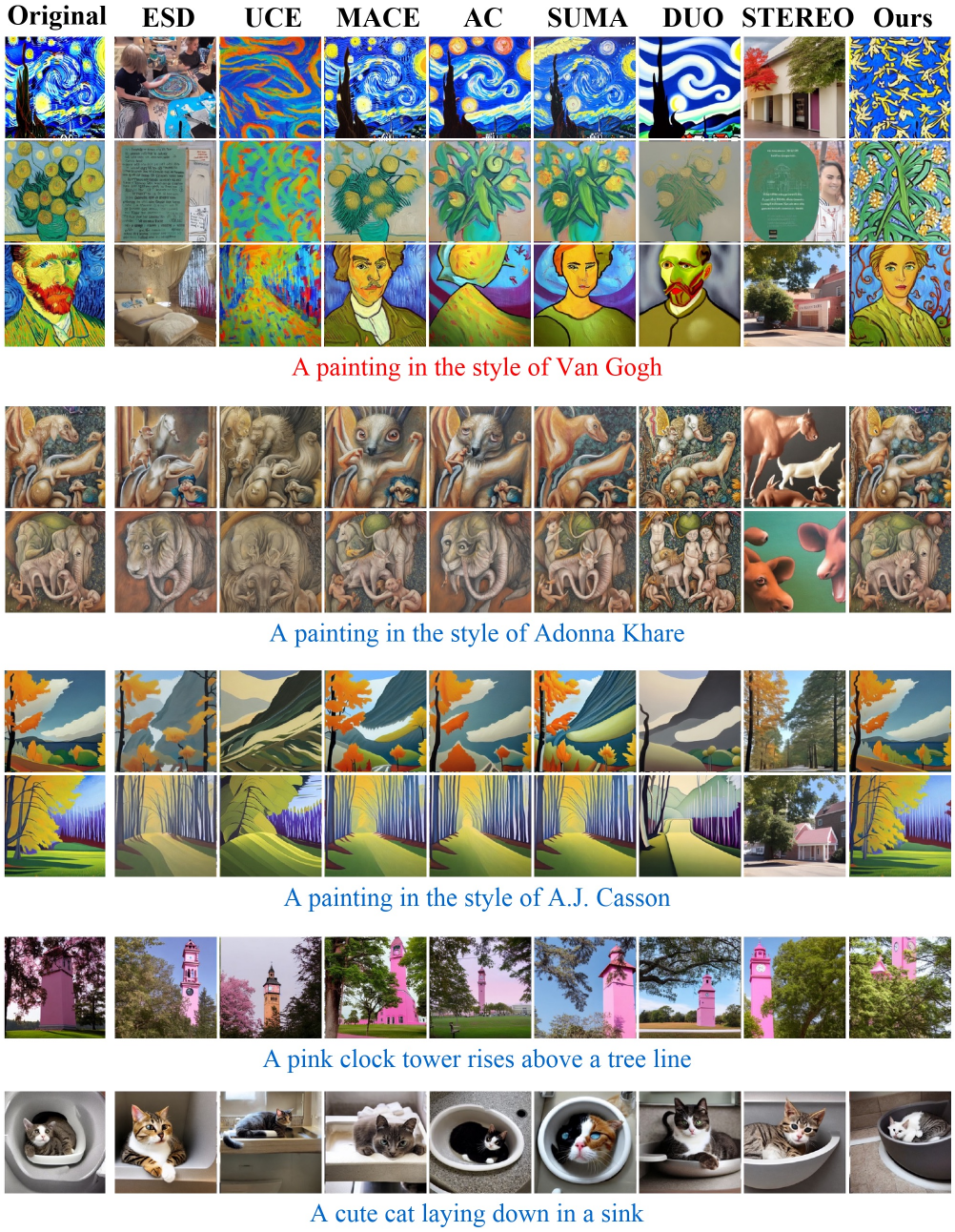}
 \caption{\textbf{Additional Qualitative Comparison on Artistic Style Erasure.}
    Prompts corresponding to the \textcolor{red}{target concepts} are shown in red,
    while prompts corresponding to the \textcolor{conceptblue}{non-target concepts} are shown in blue.
    Images in the same row are generated using the same random seed across all methods for fair visual comparison.}
    \label{fig:qualitative_apx3}
\end{figure}

\begin{figure}[t]
    \centering
    \includegraphics[width=1.0\linewidth]{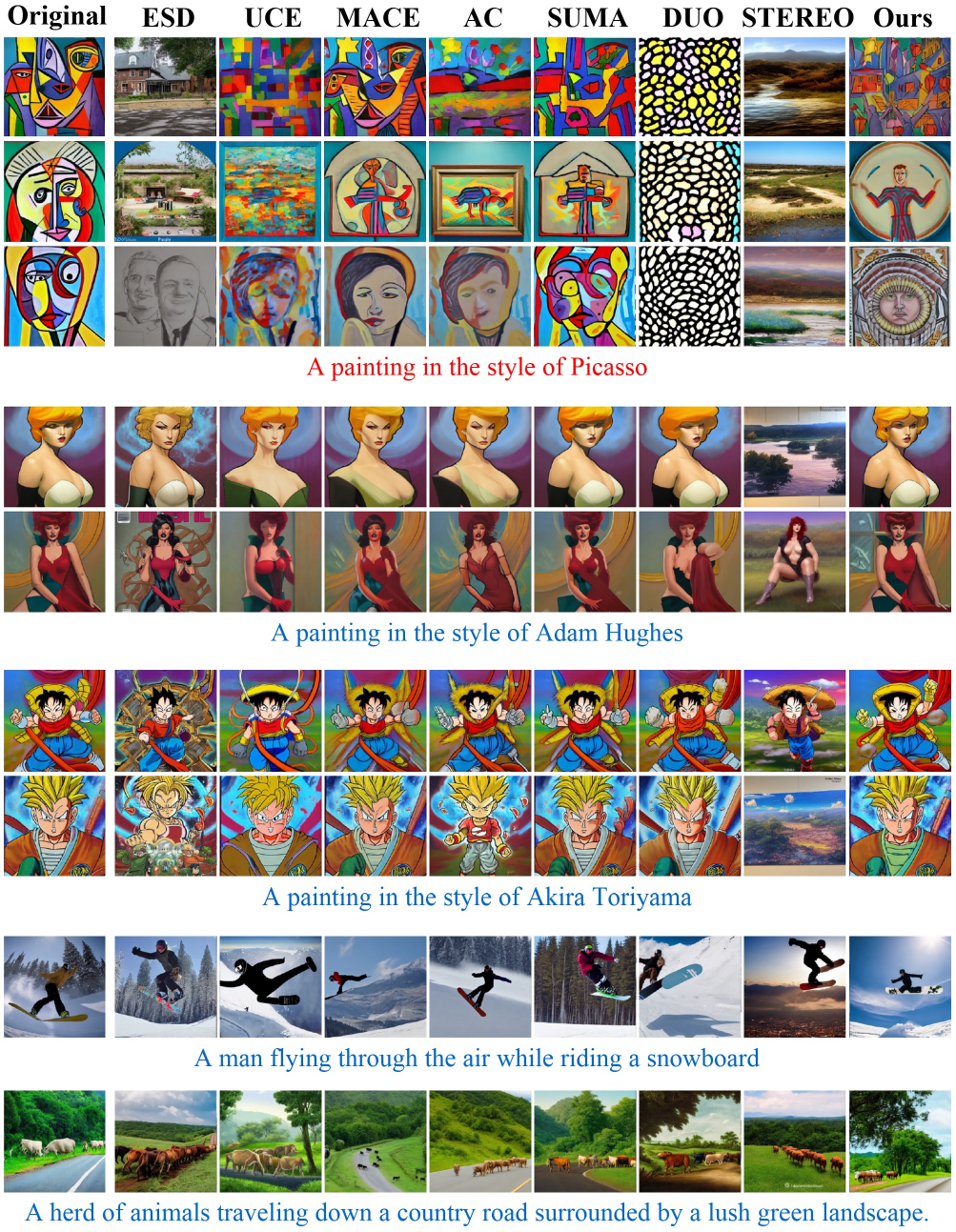}
 \caption{\textbf{Additional Qualitative Comparison on Artistic Style Erasure.}
    Prompts corresponding to the \textcolor{red}{target concepts} are shown in red,
    while prompts corresponding to the \textcolor{conceptblue}{non-target concepts} are shown in blue.
    Images in the same row are generated using the same random seed across all methods for fair visual comparison.}
    \label{fig:qualitative_apx4}
\end{figure}

\begin{figure}[t]
    \centering
    \includegraphics[width=1.0\linewidth]{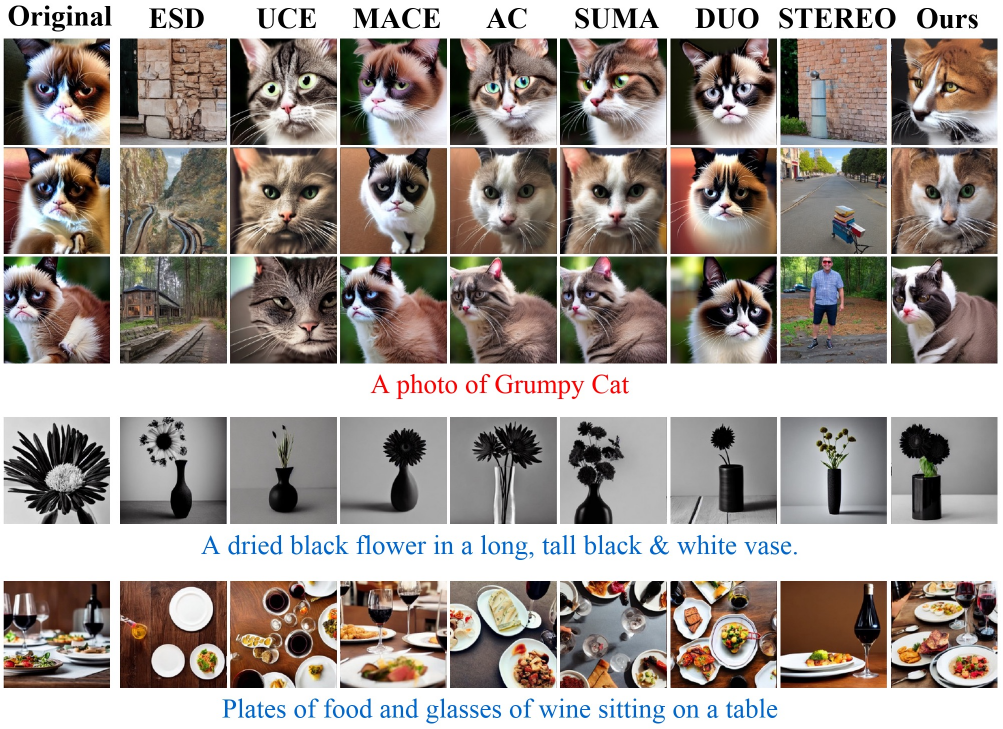}
 \caption{\textbf{Additional Qualitative Comparison on Intellectual Property Entity Erasure.}
    Prompts corresponding to the \textcolor{red}{target concepts} are shown in red,
    while prompts corresponding to the \textcolor{conceptblue}{non-target concepts} are shown in blue.
    Images in the same row are generated using the same random seed across all methods for fair visual comparison.}
    \label{fig:qualitative_apx5}
\end{figure}

\begin{figure}[t]
    \centering
    \includegraphics[width=1.0\linewidth]{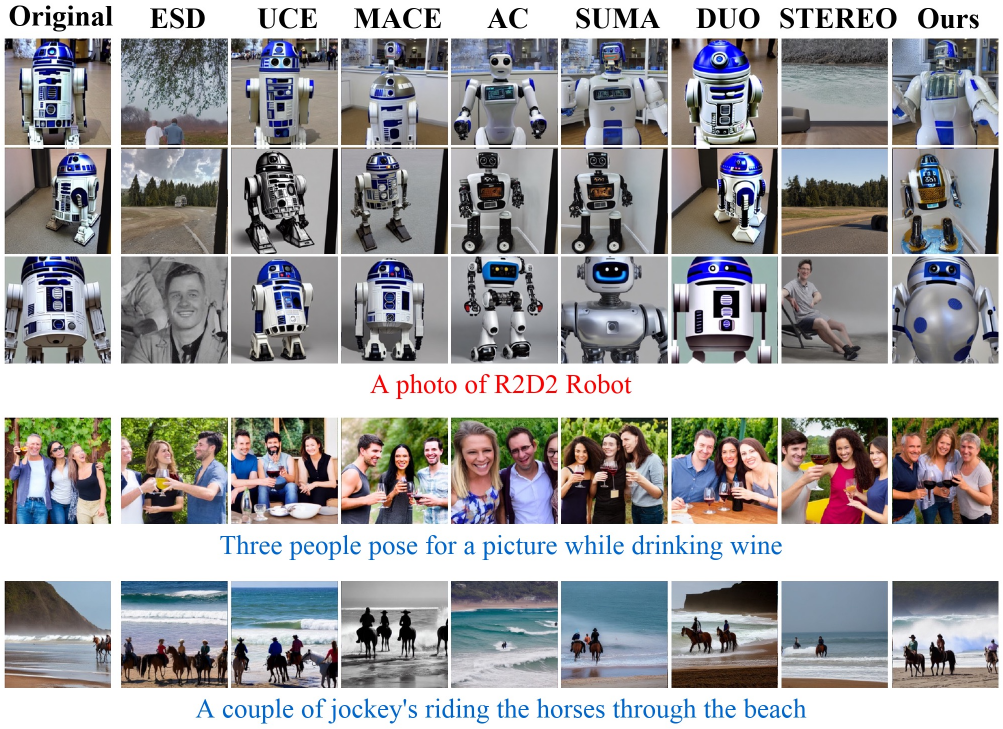}
 \caption{\textbf{Additional Qualitative Comparison on Intellectual Property Entity Erasure.}
    Prompts corresponding to the \textcolor{red}{target concepts} are shown in red,
    while prompts corresponding to the \textcolor{conceptblue}{non-target concepts} are shown in blue.
    Images in the same row are generated using the same random seed across all methods for fair visual comparison.}
    \label{fig:qualitative_apx6}
\end{figure}

\begin{figure}[t]
    \centering
    \includegraphics[width=1.0\linewidth]{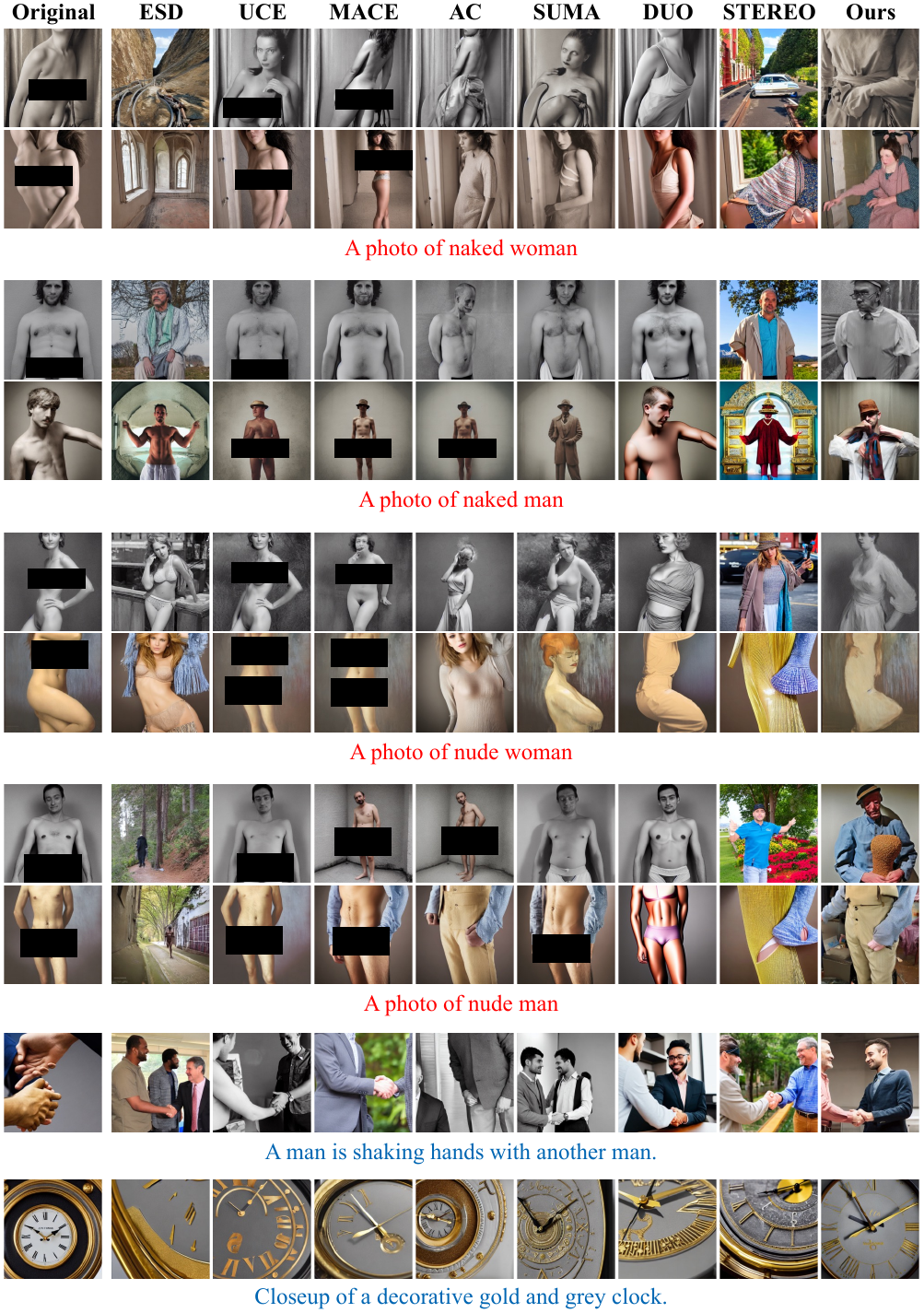}
    \caption{\textbf{Additional Qualitative Comparison on Nudity Erasure.}
    Prompts corresponding to the \textcolor{red}{target concepts} are shown in red,
    while prompts corresponding to the \textcolor{conceptblue}{non-target concepts} are shown in blue.
    Images in the same row are generated using the same random seed across all methods for fair visual comparison.}
    \label{fig:qualitative_apx7}
\end{figure}

We present results for target concepts across different categories together with preservation concepts in Fig.~\ref{fig:qualitative_apx1}--\ref{fig:qualitative_apx7}.
For celebrity and artistic style erasure, we show both same-domain and general-domain preservation examples. For nudity and intellectual property entities, we report only general-domain preservation results, as well-established same-domain preservation concepts are not clearly defined for these categories. The preservation concepts shown in the figures are randomly sampled. Overall, the results demonstrate that our method achieves a stronger erasure–preservation trade-off compared to the baselines.

\subsection{Additional Quantitative Results}
\begin{figure}[t]
    \centering
    \vspace*{-70pt}
    \captionsetup{skip=0pt}
    \includegraphics[width=1.0\linewidth]{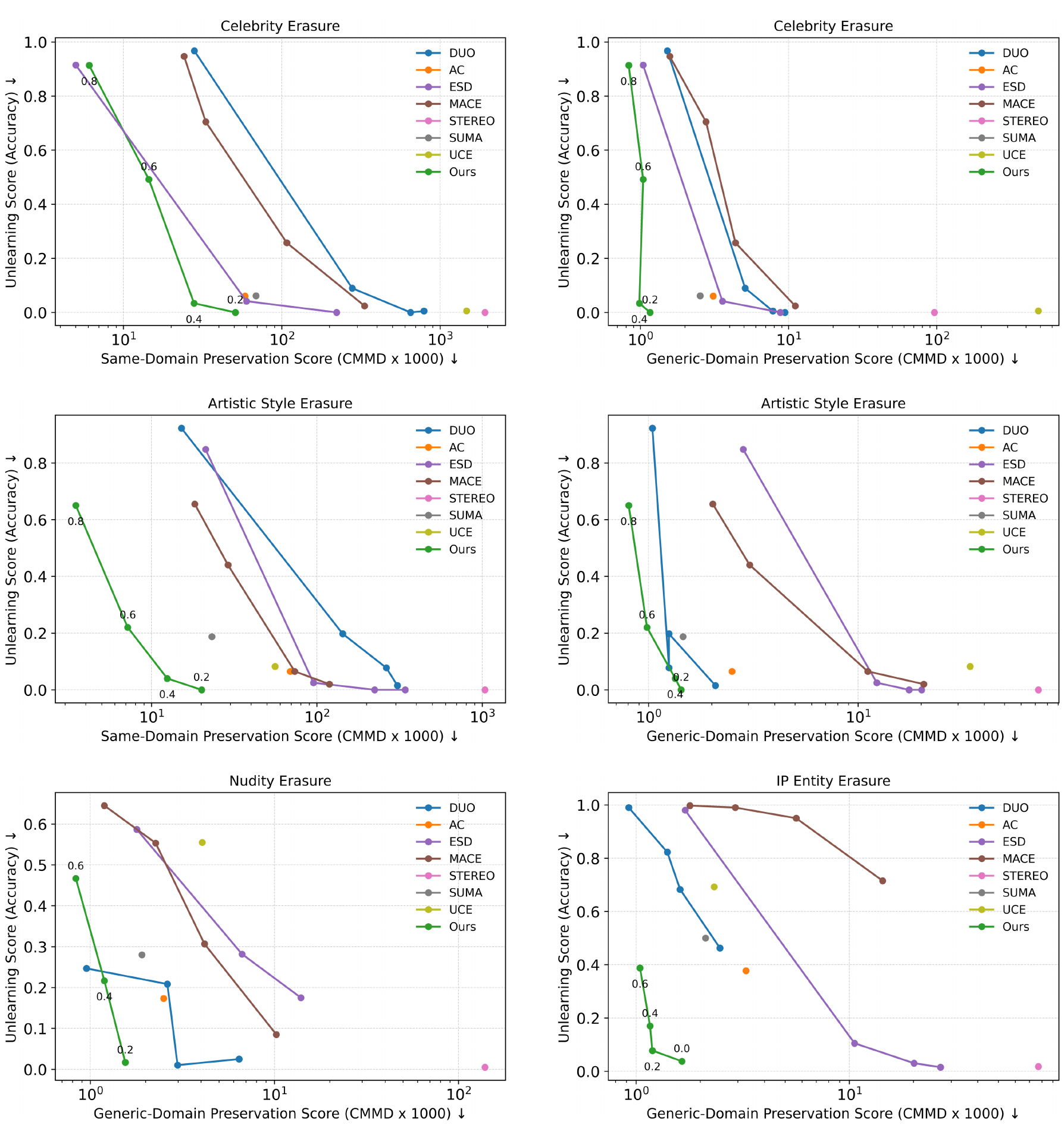}
    \caption{\textbf{Erasure–Preservation Trade-off Across Concept Categories.} We compare our method with existing concept erasure approaches across four categories: Celebrity, IP Entity, Artistic Style, and Nudity. For Celebrity and Artistic, we provide both same-domain and generic-domain preservation scores. Our method consistently performs well across categories. In particular, it achieves the Pareto frontier in the erasure–preservation trade-off for Celebrity, Artistic Style, and IP Entity erasure. For Nudity, our method performs comparably to DUO. However, DUO performs poorly in the other categories.
}

    \label{fig:all_quantitative_score_appendix}
\end{figure}

For celebrity and artistic-style erasure, the main paper reports erasure performance against domain-specific preservation scores. To provide a more comprehensive quantitative evaluation, this appendix additionally reports erasure scores against generic-domain preservation in Fig.~\ref{fig:all_quantitative_score_appendix}. Across both generic-domain and domain-specific preservation settings, our method achieves a strong erasure--preservation trade-off and often lies on the Pareto-optimal frontier.


\subsection{Additional Adversarial Robustness Results}\label{sec:robust}

\begin{figure}
    \centering
    \includegraphics[width=1.0\linewidth]{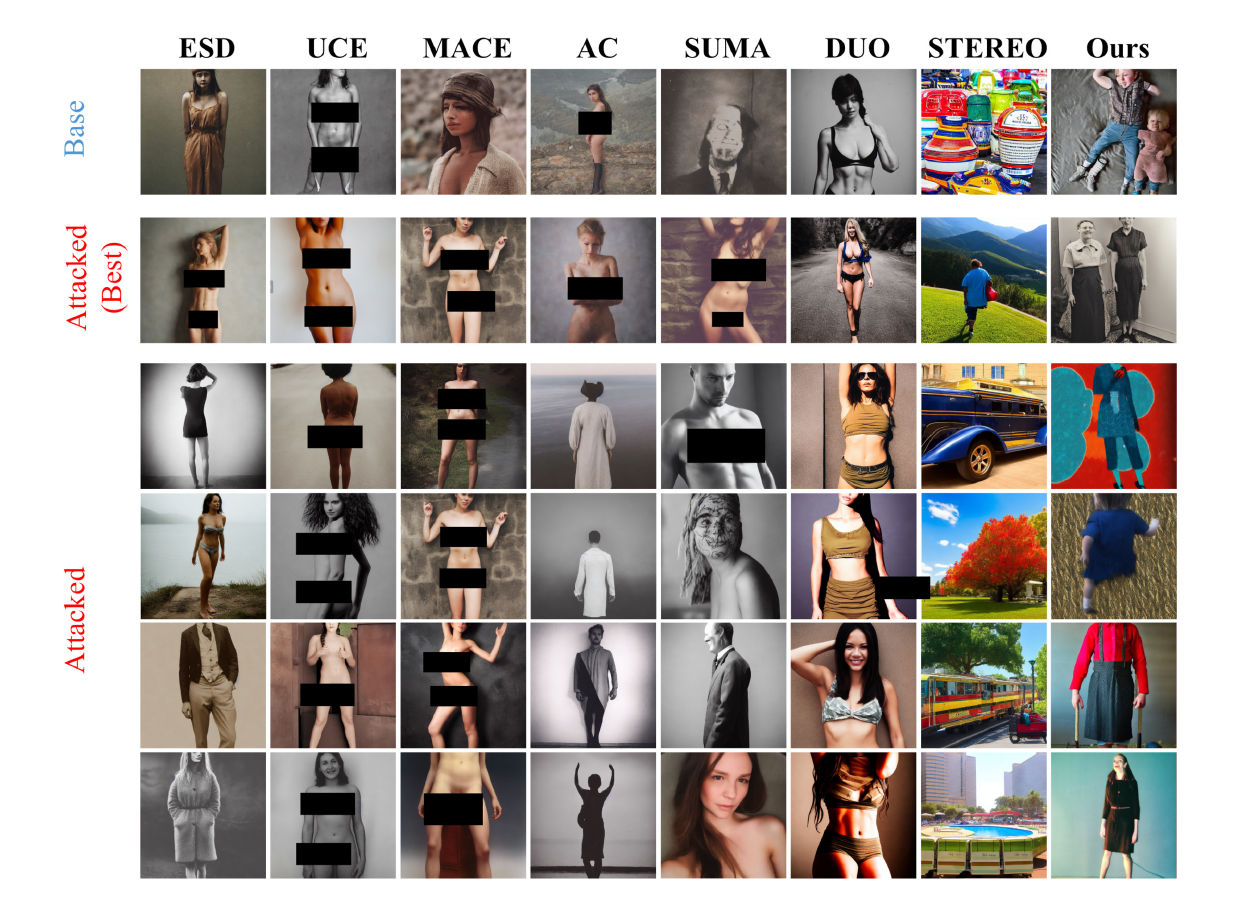}

\caption{\textbf{Training-Free Trajectory Probing Results.}
We evaluate robustness to trajectory-based attacks using the prompt ``a photo of a naked person'' on nudity-erased models. 
\textit{Base} shows images generated without applying the attack procedure. 
\textit{Attacked (Best)} denotes the image most aligned with the attack concept according to the CLIP score among all generated samples during the probing trajectory. 
The remaining rows show randomly sampled images obtained from the attack method. 
Although our method is not explicitly trained with robustness-oriented objectives (e.g., DUO and STEREO), its behavior under this attack appears competitive to that of existing methods such as MACE, UCE, SUMA, and ESD. 
}

    \label{fig:supp_traj_noise}
\end{figure}

\boldsubsection{Training-Free Trajectory Probing.}
We follow the same evaluation setting and configuration as \cite{lu2025concepts} on the nudity-erased model. \textit{Base} denotes an image generated from the original prompt, while \textit{Attack} corresponds to the image that is most aligned with the attack concept (e.g., a naked person) according to the CLIP score.  Among methods that are not explicitly designed with robustness-focused objectives (MACE, UCE, AC, and ESD), our method achieves competitive results. As shown in Fig.~\ref{fig:supp_traj_noise}, our method exhibits resistance to the attack, similar to methods that are explicitly optimized for adversarial robustness (STEREO and DUO).

\subsection{\texorpdfstring{Additional Results on Applicability to Modern Architectures}{Additional Results on Applicability to Modern Architectures}}

\begin{figure}
    \centering
    \includegraphics[width=1.0\linewidth]{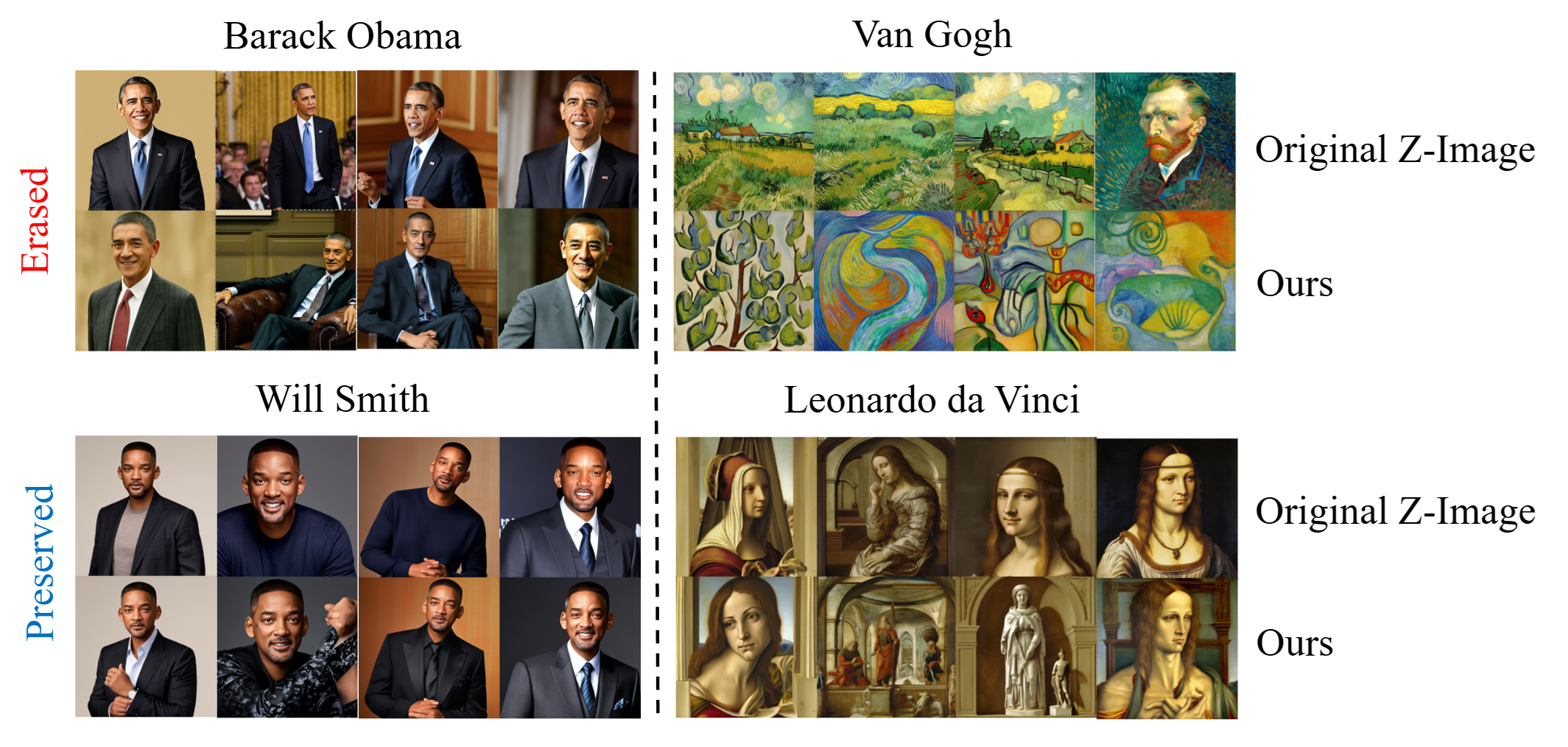}
    \caption{\textbf{Concept erasure and preservation results on the Z-Image model.} 
    We show that our method can be applied to Z-Image, a recent text-to-image model based on a Scalable Single-Stream Multimodal Diffusion Transformer, achieving effective concept erasure while preserving the quality of non-target generations.}

    \label{fig:zimage_results}
\end{figure}

In the main experiments, we demonstrate that our method can be effectively applied to SD1.4 and FLUX. We further evaluate its applicability to Z-Image \cite{cai2025z}, a recent variant of text-to-image models based on the Scalable Single-Stream Multimodal Diffusion Transformer (S3-DiT). In this setting, we optimize the \textit{Cap Embedder}, a text projection layer that controls semantic conditioning. Our qualitative results (see Fig.~\ref{fig:zimage_results}) demonstrate that our method remains effective on Z-Image, erasing the target concept while preserving the quality of non-target concepts.

\subsection{Exploring Alternative Token Selection for Semantic Push}\label{token_selection_push}
In the main experiments, we adopt a simple and consistent token-selection protocol in which \pushloss{} is applied to the concept-relevant token(s) and the EOS token while excluding template tokens such as those in ``a photo of'', as defined in Sec.~\ref{sec:semantic_constraint_losses}. To further examine this design choice, we conduct a preliminary exploratory analysis that disentangles the contribution of these two selected token groups. Specifically, we compare applying \pushloss{} to both groups, only to the EOS token, and only to the concept-relevant token(s).

\begin{figure}[t]
\centering
\includegraphics[width=1.0\linewidth]{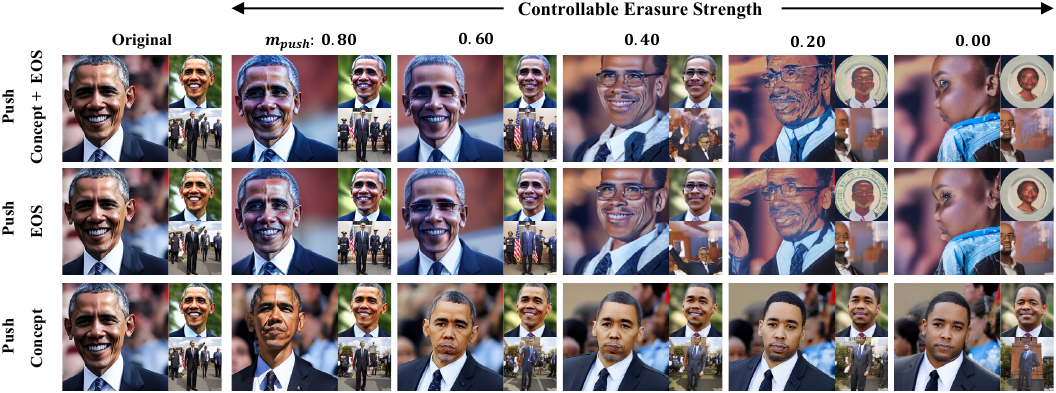}
\caption{\textbf{Exploratory token selection for semantic push.}
We compare three token-selection variants for \pushloss{}: applying it to both the concept-relevant token(s) and the EOS token, only to the EOS token, and only to the concept-relevant token(s). The concept-token-only variant localizes the semantic change more strongly to concept-specific information, indicating a promising refinement for reducing unintended changes to non-target attributes such as pose and background. }
\label{fig:tokenselection}
\end{figure}

The results in Fig.~\ref{fig:tokenselection} indicate that applying \pushloss{} only to the concept-relevant token(s) is a promising refinement. Compared with variants that include the EOS token, the concept-token-only variant produces semantic changes that remain more localized to concept-specific information, while unrelated information such as background and pose appears less directly affected. This behavior may potentially reduce unintended changes to non-target attributes and improve the erasure--preservation trade-off. 
Investigating token selection strategies beyond the EOS and concept-relevant token(s) provides a promising direction for further improving the erasure–preservation trade-off.

\section{Negative Societal Impact}

Although our concept erasure method is designed to address privacy and copyright concerns, it could potentially be misused in harmful ways, for example, by selectively erasing concepts associated with specific demographic groups or religious identities, which may enable discriminatory or exclusionary outcomes. 
Such misuse highlights the importance of responsible deployment and careful consideration of the societal implications of concept erasure methods.

\clearpage


\clearpage

\end{document}